%% file: ijcai16.tex
\documentclass{article}
\usepackage{ijcai16}
\usepackage{graphics}
\usepackage{graphicx}
\usepackage{subfigure}
\usepackage{times}
\usepackage{helvet}
\usepackage{courier}
\usepackage{mathrsfs}
\usepackage{amsmath, amsthm, amssymb, multirow, paralist, mathtools}
\usepackage{algorithm,algorithmic}
\usepackage{tikz}
\usepackage{pgfplots}
\usepackage{pgfplotstable}
\usepackage{bm}
\pgfplotsset{compat=newest}
\usetikzlibrary{bayesnet}
\def \bo {\mathbf{o}}
\def \bbm {\mathbf{m}}
\def \bI {\mathbf{I}}
\def \bW {\mathbf{W}}

\def \bZ {\mathbf{Z}}
\def \bz {\mathbf{z}}
\def \bu {\mathbf{u}}
\def \bv {\mathbf{v}}
\def \bU {\mathbf{U}}
\def \bV {\mathbf{V}}
\def \bR {\mathbf{R}}
\def \bx {\mathbf{x}}
\def \bw {\mathbf{w}}
\def \btheta {\bm\theta}
\def \bTheta {\mathbf{\Theta}}
\def \bphi {\bm\phi}
\def \bPhi {\mathbf{\Phi}}

\usepackage{times}

\title{Online Bayesian Collaborative Topic Regression}
\author{Chenghao Liu$^{1,2}$, Tao Jin$^1$, Steven C.H. Hoi$^2$, Peilin Zhao$^3$, Jianling Sun$^1$\\
$^1$School of Computer Science and Technology, Zhejiang University, China\\
$^2$School of Information Systems, Singapore Management University, Singapore\\
$^3$Institute for Infocomm Research, A*STAR, Singapore\\
twinsken@zju.edu.cn, taoj@zju.edu.cn, chhoi@smu.edu.sg, zhaop@i2r.a-star.edu.sg, sunjl@zju.edu.cn
}

\begin{document}

\maketitle

\begin{abstract}
Collaborative Topic Regression (CTR) combines ideas of probabilistic matrix factorization (PMF) and topic modeling (e.g., LDA) for recommender systems, which has gained increasing successes in many applications. Despite enjoying many advantages, the existing CTR algorithms have some critical limitations. First of all, they are often designed to work in a batch learning manner, making them unsuitable to deal with streaming data or big data in real-world recommender systems. Second, the document-specific topic proportions of LDA are fed to the downstream PMF,  but not reverse, which is sub-optimal as the rating information is not exploited in discovering the low-dimensional representation of documents and thus can result in a sub-optimal representation for prediction. In this paper, we propose a novel scheme of Online Bayesian Collaborative Topic Regression (OBCTR) which is efficient and scalable for learning from data streams. Particularly, we {\it jointly} optimize the combined objective function of both PMF and LDA in an online learning fashion, in which both PMF and LDA tasks can be reinforced each other during the online learning process. Our encouraging experimental results on real-world data validate the effectiveness of the proposed method.

\end{abstract}

\section{Introduction}
Collaborative Topic Regression (CTR) has been actively explored in recent years \cite{wang2011collaborative}. Instead of purely relying on Collaboretive Filgering(CF) approaches, CTR aims to leverages content-based techniques to overcome inaccurate and unreliable predictions with traditional CF methods due to data sparsity and other challenges. More specifically, CTR combines the idea of probabilistic matrix factorization (PMF) \cite{mnih2007probabilistic} for predicting ratings, and the idea of probabilistic topic modeling, e.g., Latent Dirchelet Allocation (LDA), for analyzing the content of items towards recommendation tasks. CTR has been shown as a promising method that produces more accurate and interpretable results and has been successfully applied in many recommender systems, such as tag recommendation \cite{wang2013collaborative,lu2015content}, and social recommender systems \cite{purushotham2012collaborative,kang2013ctr}.


Despite being studied actively \cite{wang2011collaborative,wang2013collaborative}, the existing CTR techniques suffer from several critical limitations. First of all, they are often designed to work in a batch mode learning fashion, by assuming that all text contents of items as well as the rating training data are given prior to the learning tasks. During the training process, both LDA and PMF models are usually trained separately in a batch training fashion. Such an approach would suffer from a huge scalability drawback when new data (users or items) may arrive sequentially and get updated frequently in a real-world online recommender system. Second, the existing CTR approach only leverages the content information to improve the CF tasks, but not reverse. The document-specific topic proportions of LDA are fed to the downstream PMF. This two-step procedure is rather suboptimal as the the rating information is not used in discovering the low-dimensional representation of documents, which is clearly not an optimal representation for prediction as the two methods are not tightly coupled to fully exploit their potential. Our work is motivated to explore more efficient, scalable, and effective techniques to maximize the potential exploiting extremes in dealing with data streams from real-world online recommender systems.

To overcome the limitations of traditional CTR, we propose a novel scheme of Online Bayesian Collaborative Topic Regression (OBCTR), which jointly optimizes a unified objective function by combining both PMF and LDA in an online learning fashion. In contrast to the original CTR model, OBCTR is able to achieve a much tighter coupling of both PMF and LDA, where both LDA and PMF tasks influence each other naturally and gradually via the joint optimization in the online learning process. This interplay yields item representations that are more suitable for making accurate and reliable rating prediction tasks.

To the best of our knowledge, the proposed OBCTR algorithm is the first online learning algorithm for solving CTR tasks with fully joint optimization of both LDA and PMF. Our encouraging results from extensive experiments on a large real-world data set show that the proposed online learning algorithms are scalable and effective, and the OBCTR technique not only outperforms the state-of-the-art methods for rating prediction tasks but also yields more suitable latent topic proportions in topic modeling tasks.

In the following, we first review some important related work, then present a formal formulation of CTR tasks and the novel Online Bayesian Collaborative Topic Regression algorithms. After that, we conduct extensive empirical studies and compare the proposed algorithms with the existing techniques, and finally set out our conclusions of this work.


\section{Related work}
In this section, we review two groups of studies related to our work, including (1) variants of CTR models and (2) online Bayesian inference.

\textbf{Variants of CTR model:} Researchers have extended CTR models to different applications of recommender systems.  Some researchers extended CTR models by integrating with other side information. In CTR-smf \cite{purushotham2012collaborative}, authors integrated CTR with social matrix factorization models to take social correlation between users into account. In LA-CTR \cite{kang2013ctr}, they assumed that users divide their limited attention non-uniformly over other people. In HFT \cite{mcauley2013hidden}, they aligned hidden factors in product ratings with hidden topics in product reviews for product recommendations. Some researchers extended CTR to other recommendation tasks. In CSTR \cite{ding2013celebrity}, authors explored how to recommend celebrities to general users in the context of social network. In CTR-SR \cite{wang2013collaborative}, authors adapted CTR model by combining both item-tag matrix and item content information for tag recommendation tasks. There were also several works that attempted to extract latent topic proportions of text information in CTR via deep learning techniques \cite{wang2014collaborative,wang2015relational,van2013deep}. However, all of these work follow the same parameter estimation scheme as \cite{wang2011collaborative} in a batch learning mode.

\textbf{Online Bayesian Inference:} Although the classical regime of online learning is based on decision theory, much progress has been made for developing online variational Bayes \cite{hoffman2010online,hoffman2013stochastic,kingma2013auto,foulds2013stochastic}. Most of them have adopted stochastic approximation of posterior distribution by sub-sampling a given finite data set, which is unsuitable for many applications where data size is unknown in advance.

To relax this assumption, researchers in \cite{broderick2013streaming,ghahramani2000online} made streaming updates to the estimated posterior. The intuition behind this idea is that we could treat the posterior after observing $T-1$ samples as the new prior for the incoming data points. Specifically, suppose the training data $\{\bo_t\}_{t\ge 0}$ are generated i.i.d. according to a distribution $p(\bo|\bx)$ and the prior $p(\bx)$ is given. Bayes' theorem implies the posterior distribution of $\bx$ given the first $T$ samples $(T \ge 1)$ satisfies
$
p(\bx|\{\bo\}^T_{t=0}) \propto p(\bx|\{\bo\}^{T-1}_{t=0})p(\bo_T|\bx).
$
For complex models, we can use approximate inference methods to compute the posterior. For example, \cite{broderick2013streaming} explored a mean-field variational Bayes algorithm for LDA inference. In addition, \cite{mcinerney2015population} introduced the population Variational Bayes (PVB) method which combines traditional Bayesian inference with the frequentist idea of the population distribution for streaming inference. \cite{shi2014online} proposed the Online Bayesian Passive-Aggressive (BayesPA) method for max-margin Bayesian inference of online streaming data. The high scalability of the above methods motivates us to propose Online Bayesian inference for CTR models.

\begin{figure*}[tb]
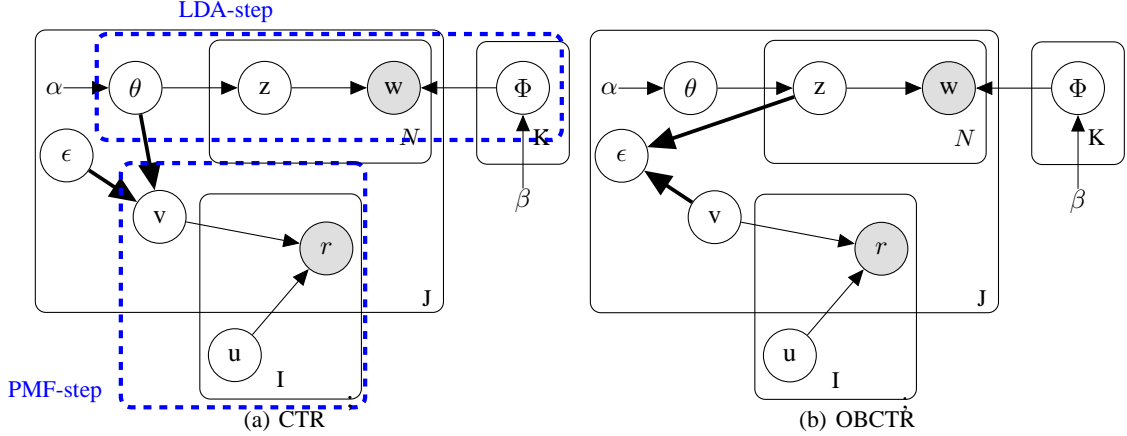

\vspace{-0.2in}
  \begin{center}
    \begin{tabular}{cc}
    \subfigure[CTR]{
    \label{fig:CTR}
    \centering
      \tikz{ %
        \node[const] (alpha) {$\alpha$} ; %
        \node[latent, right=of alpha, xshift = -0.40cm] (theta) {$\theta$} ; %
        \node[latent, left=of theta, xshift= 0.80cm, yshift = -0.90cm] (epsilon) {$\epsilon$} ;
        \node[latent, right=of theta] (z) {z} ; %
        \node[obs, right=of z] (w) {w} ; %
        \node[latent, right=of w] (Phi) {$\Phi$} ; %
        \node[const, below=of Phi,yshift =+0.02cm] (beta) {$\beta$} ; %
        \node[obs, below=of z, yshift=-0.42cm, xshift=0.82cm] (r) {$r$} ; %
        \node[latent, left=of r, yshift =0.42cm, xshift=-0.50cm] (v) {v} ; %
        \node[latent, left=of r, yshift =-1.42cm, xshift=+0.50cm] (u) {u} ; %
        \edge {alpha} {theta} ; %
        \edge[color =black, line width =0.05cm, solid] {theta} {v} ; %
        \edge[color =black, line width =0.05cm] {epsilon} {v} ; %
        \edge {z} {w} ; %
        \edge {Phi} {w} ; %
        \edge {beta} {Phi} ; %
        \edge {v} {r} ; %
        \edge {u} {r} ; %
        \edge {theta} {z} ; %
        \plate[label={[label distance=-1.5cm,shift={(0,0.5)}, text width= 4cm, text=blue]2070:LDA-step}, dashed, line width= 1.6pt, color=blue, inner sep=0.15cm, yshift=0.2cm] {plated4} {(theta) (z) (w) (Phi)};
        \plate[label={[label distance=-0.5cm, shift={(-2.5,0)},text=blue]2070:PMF-step}, dashed, line width= 1.6pt, color=blue, inner sep=0.15cm, yshift=0.2cm] {plated4} {(r) (v) (u)};
        \plate[inner sep=0.25cm] {plate3} {(Phi)} {K};
        \plate[inner sep=0.25cm, xshift=-0.12cm, yshift=0.02cm] {plate1} {(z) (w)} {$N$}; %
        \plate[inner sep=0.1cm, xshift=0.05cm, yshift=0.02cm] {plate2} {(epsilon) (v) (r) (theta) (plate1)} {J}; %
        \plate[inner sep=0.10cm, xshift=0.0cm, yshift=+0.26cm, label={[label distance=-0.5cm]2070:I}] {plate3} {(u) (r)}; %
      }
    }
    \subfigure[OBCTR]{
    \label{fig:OBCTR}
    \centering
      \tikz{ %
        \node[const] (alpha) {$\alpha$} ; %
        \node[latent, right=of alpha, xshift = -0.40cm] (theta) {$\theta$} ; %
        \node[latent, left=of theta, xshift= 0.80cm, yshift = -0.90cm] (epsilon) {$\epsilon$} ;
        \node[latent, right=of theta] (z) {z} ; %
        \node[obs, right=of z] (w) {w} ; %
        \node[latent, right=of w] (Phi) {$\Phi$} ; %
        \node[const, below=of Phi,yshift =+0.02cm] (beta) {$\beta$} ; %
        \node[obs, below=of z, yshift=-0.42cm, xshift=0.82cm] (r) {$r$} ; %
        \node[latent, left=of r, yshift =0.42cm, xshift=-0.50cm] (v) {v} ; %
        \node[latent, left=of r, yshift =-1.42cm, xshift=+0.50cm] (u) {u} ; %
        \edge {alpha} {theta} ; %
        \edge[color=black, line width =0.05cm] {v} {epsilon}; %
        \edge[color=black, line width =0.05cm] {z} {epsilon}; %
        \edge {z} {w} ; %
        \edge {Phi} {w} ; %
        \edge {beta} {Phi} ; %
        \edge {v} {r} ; %
        \edge {u} {r} ; %
        \edge {theta} {z} ; %
        \plate[inner sep=0.25cm] {plate3} {(Phi)} {K};
        \plate[inner sep=0.25cm, xshift=-0.12cm, yshift=0.02cm] {plate1} {(z) (w)} {$N$}; %
        \plate[inner sep=0.1cm, xshift=0.05cm, yshift=0.02cm] {plate2} {(epsilon) (v) (r) (theta) (plate1)} {J}; %
        \plate[inner sep=0.10cm, xshift=0.0cm, yshift=+0.26cm, label={[label distance=-0.5cm]2070:I}] {plate3} {(u) (r)}; %
      }
    }
   \end{tabular}
  \end{center}\vspace{-0.2in}
  \caption{The graphical models of CTR (left) and OBCTR (right). (a) CTR consists of two steps: (i) first runs LDA-step, and then feeds topic proportions $\btheta_j$ to the PMF-step. Note that it regards the item latent offset as $\bv_j \sim \mathcal{N}(\btheta_j,\frac{1}{\sigma_\epsilon^2}\bI_K)$ in the PMF-step. (b) OBCTR: jointly optimizing both LDA and CTR. We consider the effect of $\bm\epsilon_j \sim \mathcal{N}(\bv_j-
  \bar{\bz}_j,\frac{1}{\sigma_\epsilon^2}\bI_K)$ on both topic modeling and matrix factorization for rating prediction.}
\label{fig:OBCTR1}
\end{figure*}

\section{Collaborative Topic Regression: Revisited}
Suppose there are $I$ users and $J$ items. Each data sample is a 3-tuple $(i,j,r_{ij})$ where $i \in \{1,2,\cdots,I\}$ is the user index, $j \in \{1,2,\cdots,J\}$ is the item index and $r_{ij} \in \mathbb{R}$ is the rating value assigned to item $j$ by user $i$. We assume the rating data arrives sequentially in an online recommender system. Let $\bR$ denote the whole rating samples and the collection of $J$ items is regarded as a document set $\bW = \{\bw_j\}^{J}_{j=1}$. Let $\bZ=\{\bz_j\}^J_{j=1}$ and $\bTheta =\{\btheta_j\}^J_{j=1} $ denote all the topic assignments and topic proportions of each item. We represent users and items in a shared latent low-dimensional space of dimension $K$, which is equal to the number of topics, user i is represented by a latent vector $\bu_i \in \mathbb{R}^K$ and item j by a latent vector $\bv_j \in \mathbb{R}^K$.


Figure \ref{fig:CTR} shows the graphical model of CTR. Basically, the CTR model assumes that each item is generated by a topic model and additionally includes a latent variable $\bm\epsilon_j$ which offsets the topic proportions $\btheta_j$ when modeling the user's latent vector. This offset variable $\bm\epsilon_j$ can capture the item preference of a particular user based on their ratings. Assume there are $K$ topics $\bPhi=\{\bphi_k\}^{K}_{k=1}$. The generative process of the CTR model is as follows:
\begin{enumerate}
\item For each user $i$, draw user latent vector\\$\bu_i \sim \mathcal{N}(0,\frac{1}{\sigma_u^2}\bI_K)$
\item For each item $j$,
\begin{enumerate}
\item Draw topic proportions $\btheta_j \sim Dirichlet(\alpha)$.
\item Draw item latent offset $\bm\epsilon_j \sim \mathcal{N}(0,\frac{1}{\sigma_\epsilon^2}\bI_K))$ and set the item latent vector as $\bv_j = \bm\epsilon_j+\btheta_j$.
\item For each word $w_{jn}$($1\le n \le N_j$),
\begin{enumerate}
\item Draw topic assignment $z_{jn} \sim Mult(\btheta_j)$.
\item Draw word $w_{jn} \sim Mult(\bphi_{z_{jn}})$.
\end{enumerate}
\end{enumerate}
\item For each user-item pair $(i,j)$, draw the rating
$
r_{ij} \sim \mathcal{N}(\bu_i^T\bv_j,\frac{1}{\sigma_r^2}).
$
\end{enumerate}

In step 2 (c) ii. $\bphi_{z_{jn}}$ denotes the topic selected by the non-zero entry of $z_{jn}$. The topics are random samples drawn from a prior, e.g., $\bphi_k \sim Dirichlet(\beta)$. Note that $\bv_j = \bm\epsilon_j+\btheta_j$, where $\bm\epsilon_j \sim \mathcal{N}(0,\frac{1}{\sigma_\epsilon^2}\bI_K)$, is equivalent to $\bv_j \sim \mathcal{N}(\btheta_j,\frac{1}{\sigma_\epsilon^2}\bI_K)$. Given the document set $\bW$ and rating data $\bR$, we let $\bU=\{\bu_i\}^I_{i=1},\bV=\{\bv_j\}^J_{j=1}$, the goal of CTR is to infer the posterior distribution
\begin{eqnarray}
\label{objective1}
&p(\bU,\bV,\bZ,\bPhi,\bTheta |\bW, \bR) \propto p_0(\bU,\bV,\bZ,\bPhi,\bTheta) \nonumber\\
&p(\bW|\bZ,\bPhi)p(\bZ|\bTheta)p(\bV|\bTheta) p(\bR|\bU,\bV).
\end{eqnarray}
Because computing the full posterior of $\bU,\bV,\bZ,\bPhi,\bTheta$ directly is intractable, CTR proposed a heuristic two-stage batch learning method for approximate inference . First, CTR approximately infers posterior $p(\bZ,\bPhi,\bTheta|\bW)$ of LDA model via variational inference method \cite{blei2003latent}. Then, it applies ALS algorithm \footnote{CTR adopts the ALS algorithm \cite{hu2008collaborative} to solve an implicit feedback problem. In our context, we use the SGD algorithm \cite{koren2009matrix} since ratings data are explicit.} for learning the posterior $p(\bU,\bV|\bR, \bTheta)$ of PMF model by feeding the results of $\bTheta$ in the first step. This batch learning approach only leverages the content information to improve the CF tasks, but not reverse and tends to get trapped into local optimum.

\section{Online Bayesian Collaborative Topic Regression}
\label{OCTRbad}
Before introducing our novel online parameter estimation method of Online Bayesian Collaborative Topic Regression (OBCTR), we first modify the graphical model of CTR as shown in Figure \ref{fig:OBCTR1} to jointly optimize both LDA and PMF and adapt to our online learning method. It is worth noting that this minor modification \ref{fig:OBCTR} does not break the main structure of CTR, and our online parameter estimation method could be applied to the various variants of CTR introduced in Section 2.

CTR depicts the generative process of $\bv_j$ with $\bv_j \sim \mathcal{N}(\btheta_j,\frac{1}{\sigma_\epsilon^2}\bI_K)$. In their parameter estimation method, topic proportions $\btheta_j$ (result of LDA) provide features for $\bv_j$ in PMF, but information flow is one-way, which ignores that $\bv_j$ could provide feedback to guide the extraction of topic proportions $\btheta_j$ (they estimates $\btheta_j$ via traditional LDA algorithm which only based on $\bW$ not $\bv_j$). To address this limitation, we first assume that the item latent vector $\bv_j$ is directly close to $\bar{\bz}_j$, where $\bar{\bz}_j$ is a vector with element $\bar{\bz}_j=\frac{1}{N}\sum^N_{n=1}\mathbb{I}(z^k_n=1)$ and $\mathbb{I}$ is the indicator function that equals to 1 if predicate holds otherwise 0. In this way, $\bv_j$ can directly influence topic assignments $\bz_j$ during the procedure of inferring LDA model (variable $\bz_j$ plays a key role in LDA since other hidden variable $\bPhi$ and $\bTheta$ depend on $\bz_j$ and we can easily derive the update rule of them based on $\bz_j$). Second, we replace the generative process of $p(\bv_j|\btheta_j)$ with $p(\epsilon_j|\bar{\bz}_j,\bv_j)$, $\bm\epsilon_j \sim \mathcal{N}(\bv_j-\bar{\bz}_j,\frac{1}{\sigma_\epsilon^2}\bI_K)$, as shown in Figure \ref{fig:OBCTR1}. In our setting, $\bv_j$ and $\bar{\bz}_j$ are conditionally dependent, which means their probability of occurrence depends on either event's occurrence and allows two-way interation. In addition, instead of learning two point estimates of coefficients $\bu_i,\bv_j$, we take a more general Bayesian-style approach and learn the posterior distribution $q(\bu_i,\bv_j)$ in an online method. For rating prediction, we take a weighted average over all the possible latent vectors $\bu_i$ and $\bv_j$, or more precisely, an expectation of the prediction over $q(\bu_i,\bv_j)$ which is defined as $\hat{r}_{ij}\triangleq \mathbb{E}[\bu_i^\top\bv_j]$.

Finally, Algorithm \ref{OBCTR} summarizes the detailed framework of the proposed OBCTR algorithm. At each round t, we receive data sample and update both the parameters of LDA and PMF. The following discusses the optimization and each step of the algorithm in detail.

\begin{algorithm}
\caption{The Online Bayesian CTR (\textbf{OBCTR})}
\label{OBCTR}
\begin{algorithmic}
\STATE {\textbf{Initialize} $\bU,\bV,\bZ$ randomly.}
\FOR {t = 1 \TO $\infty$}
\STATE{Receive data sample $(i,j,r_{ij},\bw_j)$}
\STATE{Draw samples $\bz_j^t$ from Eq. (\ref{updatez})}
\STATE{Discard $B$ burn-in sweeps, use the rest samples to update $\bu_i,\bv_j,\bPhi$ following Eq. (\ref{updateu}),(\ref{updatev}),(\ref{updatePhi})}
\ENDFOR
\STATE {\textbf{Output:} $\bU,\bV$ and $\bZ$}
\end{algorithmic}
\end{algorithm}

Now, we propose our novel online parameter estimation method of Online Bayesian Collaborative Topic Regression (OBCTR) which is efficient and scalable for learning from data streams. Let us first review the objective function of CTR defined in (\ref{objective1}), from a variational point of view, this posterior is identical to the solution of the following optimization problem:
\begin{align}
\label{objective}
&\min\limits_{q(\bU,\bV,\bZ,\bPhi,\bTheta)} KL[q(\bU,\bV,\bZ,\bPhi,\bTheta)\|p_0(\bU,\bV,\bZ,\bPhi,\bTheta))] \nonumber\\
&\quad\quad\quad\quad\quad-\mathbb{E}_q[\log p(\bW|\bZ,\bPhi)p(\bV|\bTheta) p(\bR|\bU,\bV)] \nonumber\\
&\quad\qquad s.t.\quad q(\bU,\bV,\bZ,\bPhi,\bTheta) \in \mathcal{P},
\end{align}
where $KL(q\|p)$ is the Kullback-Leibler divergence, and $\mathcal{P}$ is the space of probability distributions. If we add the constant $\log p(\bW)p(\bR)$ to the objective, it is the minimization of $KL(q(\bU,\bV,\bZ,\bPhi,\bTheta)\|p(\bU,\bV,\bZ,\bPhi,\bTheta|\bW,\bR))$, which is similar with the variational formulation of original LDA \cite{blei2003latent}.
Formally, we formulate our OBCTR model as the optimization problem below:
\begin{align}
\label{objective}
&\min\limits_{q(\bU,\bV,\bZ,\bPhi,\bTheta)} KL[q(\bU,\bV,\bZ,\bPhi,\bTheta)\|p_0(\bU,\bV,\bZ,\bPhi,\bTheta))] \nonumber\\
&\quad\quad\quad\quad\quad-\mathbb{E}_q[\log p(\bW|\bZ,\bPhi)p(\bm\epsilon|\bZ,\bV) p(\bR|\bU,\bV)] \nonumber \\
&\qquad\qquad s.t.\quad q(\bU,\bV,\bZ,\bPhi,\bTheta) \in \mathcal{P},
\end{align}
where $p(\bm\epsilon|\bZ,\bTheta)=\prod^J_{j=1}p(\bm\epsilon_j|\bar{\bz}_j,\bv_j)$. Inspired by streaming Bayesian inference \cite{broderick2013streaming,ghahramani2000online}, on the arrival of new data $(i,j,r_{ij},\bw_j)$, if we treat the posterior after observing $t-1$ samples as the new prior, the post-data posterior distribution $q_{t+1}(\bu_i,\bv_j,\bz_j,\bPhi,\bTheta)$ is equivalent to the solution of the following optimization problem:
\begin{align}
\label{onlineobjective}
&\min\limits_{q} KL[q(\bu_i,\bv_j,\bz_j,\bPhi,\bTheta)\|q_t(\bu_i,\bv_j,\bz_j,\bPhi,\bTheta))] \nonumber \\
&\quad\quad-\mathbb{E}_q[\log p(\bw_j|\bz_j,\bPhi) p(\bm\epsilon_j|\bar{\bz}_j,\bv_j)p(r_{ij}|\bu_i^\top\bv_j)] \nonumber \\
&s.t. \quad q(\bu_i,\bv_j,\bz_j,\bPhi,\bTheta) \in \mathcal{P}.
\end{align}
This problem is intractable to compute. With the mean field assumption that $q(\bu_i,\bv_j,\bz_j)=q(\bu_i)q(\bv_j)q(\bz_j)$, we can solve this problem via an iterative procedure that alternatively updates each factor distribution as follows in detail.

\textbf{For $\bu_i$:} By fixing the distribution $q(\bv_j)$, we can ignore irrelevant terms and solve
\begin{align}
\min\limits_{q(\bu_i)} KL[q(\bu_i)q(\bv_j)\|q_t(\bu_i)p(r_{ij}|\bu_i^\top\bv_j)]. \nonumber
\end{align}
The optimal solution has the following closed form solution:
\begin{align}
q_{t+1}(\bu_i)\propto q_t(\bu_i)\exp(\mathbb{E}_{q(\bv_j)}[\log p(r_{ij} |\bu_i^\top\bv_j)]). \nonumber
\end{align}
If initial prior is normal $q_0(\bu_i)=\mathcal{N}(\bu_i;\bbm_{ui}^0,\Sigma_{ui}^0)
$, by induction we can show that the inferred distribution at each round is also a normal distribution. Let us assume $q_t(\bu_i)= \mathcal{N}(\bu_i; \bbm_{ui}^t,\Sigma_{ui}^t)$. Then, we have
\begin{small}
\begin{eqnarray}
q_{t+1}(\bu_i) \hspace{-0.25in}&& \propto \exp(-\frac{1}{2}(\bu_i-\bbm_{ui}^t)^\top(\Sigma_{ui}^t)^{-1}(\bu_i-\bbm_{ui}^t) \nonumber\\
&&\hspace{0.15in}+\mathbb{E}_{q(\bv_j)}[-\frac{(r_{i,j}-\bu_i^\top\bv_j)^2}{2\sigma_r^2}]) \nonumber \\
&& = \mathcal{N}(\bu_i;\bbm_{ui}^\ast,\Sigma_{ui}^\ast),\nonumber
\end{eqnarray}
\end{small}
where the posterior parameters are computed as
\begin{small}
\begin{eqnarray}
\label{updateu}
&&\Sigma^\ast_{ui} = ((\Sigma^t_{ui} )^{-1}+\frac{\bbm_{vj}\bbm_{vj}^
\top}{\sigma_r^2\bI_K})^{-1}, \\
&&\bbm^\ast_{ui}  =  \bbm^t_{ui} + \frac{r_{i,j}-\bbm_{vj}^\top\bbm^t_{ui}}{\sigma_r^2+\bbm_{vj}^\top\Sigma^t_{ui}\bbm_{vj}}\Sigma^t_{ui}\bbm_{vj}.\nonumber
\end{eqnarray}
\end{small}
To make it more efficient, we only update the diagonals of covariance matrix $\Sigma_{ui}^\ast$.

\textbf{For $\bv_j$:} The update rule of $\bv_j$ is similar to $\bu_i$ except adding a Gaussian distribution $p(\bm\epsilon_j|\bar{\bz}_j,\bv_j)$, a constraint about the distance between $\bv_j$ and $\bar{\bz}_j$, that explains the difference between topic assignments in content and item preference based on ratings. By fixing the distribution of $q(\bu_i)$ and $q(\bz_j)$, we have the update rule
\begin{small}
\begin{eqnarray}
\hspace{-0.2in}&&\hspace{-0.2in}q_{t+1}(\bv_j)\propto q_t(\bv_j)\exp(\mathbb{E}_{q(\bu_i,\bz_j)} [\log p(r_{ij}|\bu_i^\top\bv_j) p(\bm\epsilon_j|\bar{\bz}_j,\bv_j)]) \nonumber \\
\hspace{-0.2in}&&\hspace{-0.2in} \propto \exp(-\frac{1}{2}(\bv_j-\bbm_{vj}^t)^\top(\Sigma_{vj}^t)^{-1}(\bv_j-\bbm_{vj}^t) \nonumber\\
\hspace{-0.2in}&&\hspace{-0.2in}+\mathbb{E}_{q(\bu_i)q(\bz_j)}[-\frac{(r_{i,j}-\bu_i^\top\bv_j)^2}{2\sigma_r^2}\hspace{-0.03in}-\hspace{-0.03in}\frac{(\bar{\bz}_j-\bv_j)^\top(\bar{\bz}_j-\bv_j)}{\sigma^2_\epsilon\bI_K}])\nonumber \\
\hspace{-0.2in}&&\hspace{-0.2in}= \mathcal{N}(\bv_j;\bbm_{vj}^\ast,\Sigma_{vj}^\ast),\nonumber
\end{eqnarray}
\end{small}
where the posterior parameters are computed as
\begin{small}
\begin{eqnarray}
\label{updatev}
&&\hspace{-0.25in}\Sigma_{mix}=(\Sigma_{vj}^{-1}+\frac{1}{\sigma_\epsilon^2})^{-1}, \\
&&\hspace{-0.25in}\Sigma^\ast_{vj} = ((\Sigma^t_{vj} )^{-1}+\frac{1}{\sigma^2_\epsilon\bI_K} + \frac{\bbm_{ui}\bbm_{ui}^
\top}{\sigma_r^2\bI_K})^{-1}, \nonumber\\
&&\hspace{-0.25in}\bbm_{vj}^\ast = \Sigma_{mix}\Sigma_{vj}^{-1}\bbm_{vj}^t + \Sigma_{mix}\frac{1}{\sigma_\epsilon^2}\bar{\bz}_j-\Sigma_{mix}\frac{1}{\sigma_r^2}\bbm_{ui}\nonumber\\ &&\hspace{0.15in}(\frac{\bbm_{ui}^\top\Sigma_{mix}\Sigma_{vj}^{-1}\bbm_{vj}^t + \bbm_{ui}^\top\Sigma_{mix}\frac{1}{\sigma_\epsilon^2}\bar{\bz}_j-r_{ij}
}{1+\bbm_{ui}^\top\Sigma_{mix}\frac{1}{\sigma_r^2}\bbm_{ui}}). \nonumber
\end{eqnarray}
\end{small}
Besides, we adopt the same strategy that only updating the diagonals of covariance matrix $\Sigma_{vj}^\ast$.

\textbf{For $\bPhi$ and $\bTheta$:} By fixing the distribution $q(\bZ)$, the update rule for Dirichlet distribution $\bPhi$ and $\bTheta$ is similar to the original LDA, that is,
\begin{small}
\begin{eqnarray}
\label{updatePhi}
\theta_{jk} = \frac{C_j^k+\alpha}{\sum^K_{k=1}C_j^k+K\alpha},  \quad \Phi_{kw} = \frac{C^w_k+\beta}{\sum^{D}_{w=1}C^w_k+D\beta},
\end{eqnarray}
\end{small}
where $D$ is the vocabulary size, $C_j^k$ is the number of times that terms being associated with topic k within the j-th item, $C^w_k$ is the number of times the term $w(1\le w \le D)$ being assigned to topic $k$ over the whole corpus.

\textbf{For $\bz_j$:} Given the distribution of other variables, the conditional distribution of $\bz_j$ is:
\begin{small}
\begin{eqnarray*}
&&\hspace{-0.35in}q_{t+1}(\bz_{j}|\bv_j,\bPhi,\bw_j)\nonumber\\
&&\hspace{-0.35in}\propto q_t(\bz_{j})\exp( \mathbb{E}_{q(\bPhi)q(\bv_j)}[\log p(\bw_j|\bz_{j},\bPhi)p(\bm\epsilon_j|\bar{\bz}_j,\bv_j)]) \nonumber \\
&& \hspace{-0.35in}\propto q_t(\bz_t)\exp(\sum_{n \in [N_j]}\Lambda_{z_{jn},w_{jn}}-\mathbb{E}_{q(\bv_j)}[\frac{(\bv_j-\bar{\bz}_j)^\top(\bv_j-\bar{\bz}_j)}{\sigma^2_\epsilon\bI_K}])\quad
\end{eqnarray*}
\end{small}
where $\Lambda_{z_{jn},w_{jn}}= \mathbb{E}_{q(\Phi)}[\log (\bPhi_{z_{jn},w_{jn}})]$. We can do Gibbs sampling to infer $q(\bz_j)$ by canceling out common factors. This hybird strategy has shown promising performance for LDA \cite{mimno2012sparse,shi2014online}. Specifically, the conditional distribution of one varibale $z_{jn}$ (the topic assignment of the n-th word in item $j$ ) given others $\bz_{j\neg n}$ is
\begin{small}
\begin{eqnarray}
\label{updatez}
&&\hspace{-0.3in}q(z_{jn} = k | \bz_{j\neg n},\bv_j, \bPhi, w_{jn}=w)\\
&&\hspace{-0.3in}\propto (\alpha+C^{k}_{j\neg n})\exp(\Lambda_{k,w_{jn}}+\frac{1}{2\sigma_\epsilon^2N_j} (2m_{vjk}- \frac{1+2C^{k}_{j \neg n}}{N_j})),\nonumber
\end{eqnarray}
\end{small}
where $\bz_{j\neg n}$ is the topic assignments in item $j$ (except the n-th word) and $C^{k}_{j \neg n}$ is the number of words in item $j$ (except the n-th word) that are assigned to topic $k$.


\section{Experimental Results}
Our experiments were conducted on an extended MovieLens dataset, named as ``MovieLens-10M-Plot"\footnote{We will release the dataset after the paper is accepted.}, which was originated from the MovieLens 10M\footnote{http://grouplens.org/datasets/movielens/}. Specifically, the original MovieLens 10M dataset provides a total of 10,000,053 rating records for 10,681 movies (items) by 69,878 users. However, the original dataset has very limited \textit{text} content information. We enrich the dataset by collecting additional text contents for each of the movie items. Specifically, for each movie item, we first used its identifier number to find the movie listed in the IMDb\footnote{http://www.imdb.com} website, and then collected its related text of ``plot summary".
We then combine the ``plot summary" text together with each movie's title and category text given in the MovieLens-10M dataset as a text document to represent each movie. For detailed text preprocessing, we follow the same procedure as the one described in \cite{wang2011collaborative} to process text information. Finally, we form a vocabulary with 7,689 distinct words. Note that we did not consider the CiteUlike dataset \footnote{http://www.citeulike.org/faq/data.adp} as used in the previous study \cite{wang2011collaborative}, because their dataset only provides ``like" and ``dislike" preference, which is kind of implicit feedback and thus unsuitable for our regression task. By contrast, the MovieLens-10M dataset has explicit feedback with ratings ranging from 1 to 5.

 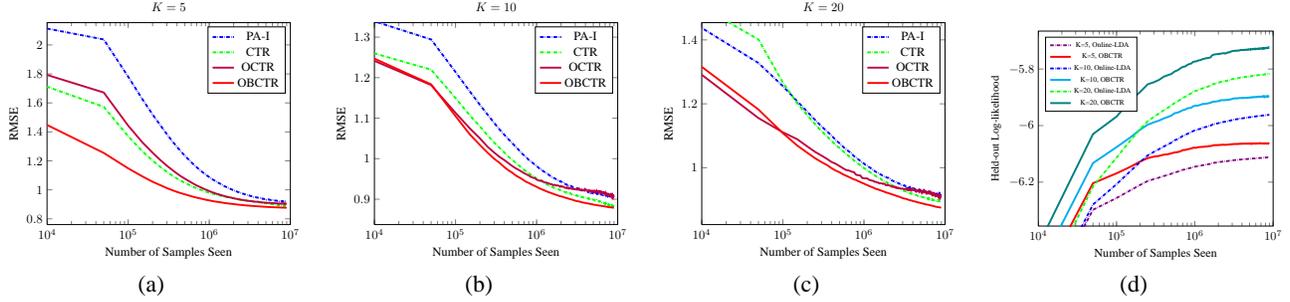
\begin{figure*}[htbp]
  \centering\vspace{-0.1in}
  \subfigure[]{
    \resizebox{1.61in}{!}{
    \input{obctr_rmse_5}
    }
    \label{fig:obctr_rmse_5}
  }
    \subfigure[]{
    \resizebox{1.61in}{!}{
    \input{obctr_rmse_10}
    }
    \label{fig:obctr_rmse_10}
  }
    \subfigure[]{
    \resizebox{1.61in}{!}{
    \input{obctr_rmse_20}
    }
    \label{fig:obctr_rmse_20}
  }
    \subfigure[]{
    \resizebox{1.61in}{!}{
    \input{likelihood}
    }
    \label{fig:likelihood}
  }
  \caption{Figure (a)(b)(c) show the evaluation of RMSE performance by different online algorithms after seeing different training data streams. Figure (d) demonstrates the online per-word predictive log likelihood comparisons between OBCTR and Online LDA}
  \label{fig:rmse}
\end{figure*}

\subsection{Experimental Setup and Metric}
The dataset has more than 10-million rating records. For each experiment, we randomly shuffle the rating records, and then divide them into two parts: the first 90\% of the shuffled rating records are used as the training data, and the rest 10\% rating data are used as test set. We also randomly draw 5\% out of the training data as the validation set for parameter selection. To make fair comparisons, all the algorithms are conducted over 5 experimental runs of different random permutations. For performance metric, we evaluate the performance of our proposed method for prediction task by measuring Root Mean Square Error (RMSE).
In the online learning experiments, we evaluate the RMSE performance on the test set after every 50,000 online iterations. In addition, we also evaluate the performance of topic modeling via the log-likelihood of each word in text collection \cite{hoffman2010online}.

\subsection{Baselines for Comparison and Experimental Settings}
In our experiments, we evaluate the proposed OBCTR algorithms for rating predictions by comparing with some important baselines as follows:
\begin{itemize}
\item \textbf{PA-I}: An online learning algorithm for solving online collaborative filtering tasks by applying the popular online Passive-Aggressive (PA) algorithm \cite{blondel2014online};
\item \textbf{CTR}: the existing Collaborative Topic Regression \cite{wang2011collaborative} . In our context, we replace the ALS algorithm \cite{hu2008collaborative} with SGD algorithm \cite{koren2009matrix} since ratings data are explicit, and keep the rest same as the original CTR (note that the LDA step is still performed in a batch manner);
\item \textbf{OCTR}: To evaluate the efficacy of joint optimization. We propose a simplified variant of OBCTR, named OCTR, which runs online LDA \cite{hoffman2010online} for LDA part and SGD for PMF part (but without joint optimization as OBCTR) sepearately. OCTR closely resembles the original CTR --- the most important difference is that we extract topic proportions from LDA part, and then feed it to the downstream update of PMF part every time data sample arrives;
\item \textbf{OBCTR}: The proposed Online Bayesian CTR algorithm in Algorithm~\ref{OBCTR}.
\end{itemize}
Besides, to evaluate the topic modeling performance, we also compare our method with the typical Online LDA method:
\begin{itemize}\vspace{-0.05in}
\item \textbf{Online-LDA}: an online Bayesian variational inference algorithom for LDA model\cite{hoffman2010online}. We take it as a baseline to evaluate how well the model fits the data with the predictive distribution.\vspace{-0.05in}
\end{itemize}
For parameter settings, we fix $\alpha=K^{-1},\beta=K^{-1}$  and find the optimal parameters for different algorithms (PA-I, CTR and OBCTR). Specifically, the parameters including $c$ in PA-I, $\sigma_u$, $\sigma_v$ and $\rho$ in CTR and OCTR, and  $\sigma_\epsilon$ and $\sigma_r$ in OBCTR. All of these parameters are found by performing a grid search as follows: $ \sigma_\epsilon,\sigma_r \in \{0.5, 1, 2, 4, 8, 16, 32\}$, $c \in \{0.01,0.1,0.2,0.5,1\}$, $\rho \in \{0.01,0.05,0.1,0.2,0.5\}$, $\sigma_u, \sigma_v \in \{0.01, 0.02, 0.04, 0.08, 0.16, 0.32\}$ and $K \in \{5,10,20\}$.
\begin{table}
\begin{tabular}{|l|l|l|l|}
\hline
K   &5  &10 &20 \\ \hline
PA-I    &\small{0.9176} \tiny{$\pm$0.0004}	&\small{0.9085} \tiny{$\pm$0.0002} &\small{0.9148} \tiny{$\pm$0.0003}	 \\ \hline
CTR     &\small{0.8874} \tiny{$\pm$0.0003}  &\small{0.8812} \tiny{$\pm$0.0005} &\small{0.8947} \tiny{$\pm$0.0007} \\ \hline
OCTR    &\small{0.9034} \tiny{$\pm$0.0006}  &\small{0.9054} \tiny{$\pm$0.0008} &\small{0.9085} \tiny{$\pm$0.0002} \\ \hline
OBCTR   &\textbf{\small{0.8763}} \tiny{$\pm$0.0006}  &\textbf{\small{0.8788}} \tiny{$\pm$0.0001} &\textbf{\small{0.8747}} \tiny{$\pm$0.0006} \\ \hline
\end{tabular}
\caption{RMSE results after a single pass over training set}\vspace{-0.2in}
\label{finalresult}
\end{table}
\subsection{Evaluation of Online Rating Prediction Tasks}
Figure \ref{fig:obctr_rmse_5},\ref{fig:obctr_rmse_10},\ref{fig:obctr_rmse_20} compares the online performance of the above methods in $K=5$, $K=10$ and $K=20$. We note that the CTR method took at least 6 hours \footnote{For the vanilla LDA inference method, a larger K value often needs more time for computation.} to precompute the parameters $\bTheta$ and $\bPhi$ by a batch variational inference algorithm. Figure \ref{fig:rmse} shows only its performance in the downstream collaborative filtering phase.

As we can see from Figure \ref{fig:obctr_rmse_5},\ref{fig:obctr_rmse_10},\ref{fig:obctr_rmse_20}, the CTR-based approaches outperform the online CF algorithm (PA-I) for most cases, which is in line the experiments in \cite{wang2011collaborative} and validates the efficacy of leveraging additional text information to improve the performance of PMF for online rating prediction tasks. Second, among different CTR-based approaches, the proposed OBCTR consistently outperforms the other algorithms for most cases. This validates the importance of jointly optimizing both online PMF and online LDA to achieve tight coupling of the two techniques. Moreover, it is interesting to find that the gap between the proposed OCTR variant and OBCTR tends to become more significant when $K$ is smaller. We conjecture that this is because when $K$ is small, the PMF performance is relatively inaccurate and thus including the joint optimization becomes more critical for enhancing the unreliable PMF prediction performance.

Finally, Table \ref{finalresult} summarizes the final test-set RMSE results after finishing the whole online learning tasks (by a single pass over the training set). Similar observations can be found , in which OBCTR achieves the lowest RMSE result on the test set for rating prediction among all the algorithms. In addition, CTR has better performance than OCTR. This is because CTR directly takes the batch LDA results (pre-computed $\bTheta$ and $\bPhi$) as input for leveraging online PMF task, while online CTR may converge relatively slowly (without the tight coupling). This again shows that it is crucial for the joint optimization in OBCTR.

\subsection{Performance on Online Topic modeling Tasks}

Figure \ref{fig:likelihood} shows the results about online average predictive log likelihood for OBCTR and Online LDA. Online learning allows us to conduct a large-scale comparison. We can see that OBCTR exhibits consistently better performance than Online LDA, which ignores ratings information, regardless of how many topics we use. That is due to the utilization of rating information to discover the low-dimensional topic proportions, where OBCTR yields additional benefit on this task.

\subsection{Evaluation of Parameter Sensitivity}
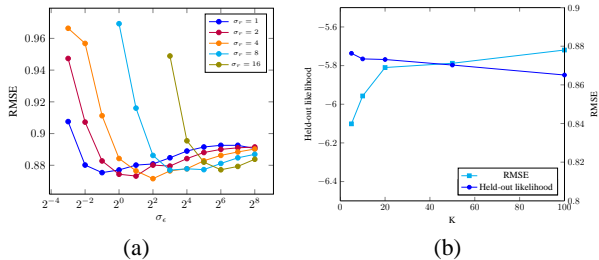
\begin{figure}[htb]
	\centering

\subfigure[]{
	\resizebox{1.45in}{!}{
		\input{parameter_vr}
	}
\label{fig:parameter}
}
\subfigure[]{
	\resizebox{1.6in}{!}{
		\input{parameter_k}
	}
\label{fig:modelsize}
}
\caption{(a) shows the evaluation of parameter influences ($\sigma_r$ and $\sigma_\epsilon$). (b) demonstrates the evaluation of OBCTR result by varying K}
	
\end{figure}

Figure \ref{fig:parameter} shows how RMSE is affected by the choice of two key parameters $\sigma_\epsilon$ and $\sigma_r$ in OBCTR.

As observed from Figure \ref{fig:parameter}, at the beginning, increasing $\sigma_\epsilon$ leads to decrease the RMSE quickly. After arriving some optimal value, increasing $\sigma_\epsilon$ further may increase the RMSE gradually. Second, we found the optimal value of $\sigma_\epsilon$ also largely depends on the setting of the parameter $\sigma_r$. When $\sigma_r$ is smaller, the optimal value of $\sigma_\epsilon$ is relatively smaller.
However, after reaching the optimal value, the further performance changing becomes limited. This indicates that overall, it is relatively easy to choose a good value of $\sigma_\epsilon$ given a fixed $\sigma_r$ setting due to its less sensitivity in the range of optimal values. Our results were consistent to the similar phenomena observed in \cite{wang2011collaborative}.


Figure \ref{fig:modelsize} demonstrates the effect of increasing model complexity $K$. This investigation is done by selecting the best achievable RMSE and log-likelihood during the grid parameter search process. As shown in the diagram, increasing the complexity of models (higher $K$ values) leads to improvement of both RMSE and log-likelihood results. However, the gain of predictive performance is paid by a significant computational overhead for more complex models (as shown in Table \ref{time}). In a practical online recommender system, one may want to choose a proper value of $K$ to balance the tradeoff between accuracy and computational efficiency.

\begin{table}[htb]
	\centering
	\begin{tabular}{|c|c|c|c|c|c|}
		\hline		
		$K$ &5	&10	&20	&50	&100 \\ \hline
		Time Ratio & 1.00	&1.29	&2.62	&5.64	&11.43 \\ \hline			
	\end{tabular}\vspace{-0.1in}
	\caption{Running time consumed for each model size ($K$). Time Ratio indicates the amount of time required compared with that of the simplest model ($K=5$). }
	\label{time}\vspace{-0.1in}
\end{table}

\section{Conclusion}

This paper investigated online learning algorithms for making Collaborative Topic Regression (CTR) techniques practical for real-world online recommender systems. Specifically, unlike CTR that loosely combines LDA and PMF, we propose a novel Online Bayesian CTR (OBCTR) algorithm which performs a joint optimization of both LDA and PMF to achieve a tight coupling. Our encouraging results showed that OBCTR converges much faster than the other competing algorithms in the online learning, and thus achieved the best prediction performance among all the compared algorithms. Our future work will analyze model interpretability and theoretical performance of the proposed algorithms.

\bibliographystyle{named}
\bibliography{ijcai16}

\end{document}

%% file: obctr_rmse_5.tex
\begin{tikzpicture}
\begin{axis}[
    xmode = log,
    xmin=1e4, xmax=1e7,  
    xlabel = {Number of Samples Seen},
    ylabel = RMSE,
    title = {$K=5$}
]
\addplot [mark=, densely dashdotted, line width=1.5pt, color=blue] table {
1	2.54329
50001	2.03788
100001	1.78084
150001	1.6249
200001	1.52014
250001	1.44488
300001	1.38443
350001	1.33588
400001	1.29632
450001	1.26301
500001	1.23545
550001	1.21171
600001	1.19057
650001	1.17198
700001	1.1557
750001	1.14159
800001	1.12824
850001	1.11646
900001	1.1068
950001	1.09633
1000001	1.08783
1050001	1.07991
1100001	1.07162
1150001	1.06575
1200001	1.05925
1250001	1.05264
1300001	1.04818
1350001	1.04351
1400001	1.03783
1450001	1.03344
1500001	1.02821
1550001	1.02415
1600001	1.02078
1650001	1.01778
1700001	1.0141
1750001	1.01053
1800001	1.00735
1850001	1.00476
1900001	1.00252
1950001	0.999341
2000001	0.997733
2050001	0.994238
2100001	0.991514
2150001	0.989802
2200001	0.987836
2250001	0.985247
2300001	0.982768
2350001	0.981662
2400001	0.979008
2450001	0.978207
2500001	0.976267
2550001	0.975284
2600001	0.973115
2650001	0.972422
2700001	0.970461
2750001	0.969355
2800001	0.967892
2850001	0.965961
2900001	0.964783
2950001	0.963725
3000001	0.962778
3050001	0.961458
3100001	0.960793
3150001	0.959397
3200001	0.958442
3250001	0.957208
3300001	0.955716
3350001	0.955866
3400001	0.95533
3450001	0.954028
3500001	0.951779
3550001	0.952151
3600001	0.951351
3650001	0.950581
3700001	0.949787
3750001	0.949168
3800001	0.947731
3850001	0.947384
3900001	0.947406
3950001	0.946541
4000001	0.945251
4050001	0.944826
4100001	0.944271
4150001	0.942869
4200001	0.942287
4250001	0.942229
4300001	0.94154
4350001	0.94039
4400001	0.939897
4450001	0.940437
4500001	0.939255
4550001	0.938078
4600001	0.937245
4650001	0.938243
4700001	0.937568
4750001	0.937435
4800001	0.936283
4850001	0.935368
4900001	0.935396
4950001	0.934415
5000001	0.933886
5050001	0.934941
5100001	0.932757
5150001	0.933491
5200001	0.933948
5250001	0.932654
5300001	0.933332
5350001	0.933019
5400001	0.933107
5450001	0.932128
5500001	0.93217
5550001	0.931233
5600001	0.931201
5650001	0.930716
5700001	0.930489
5750001	0.930584
5800001	0.929074
5850001	0.929348
5900001	0.929359
5950001	0.928155
6000001	0.928123
6050001	0.928421
6100001	0.928822
6150001	0.928486
6200001	0.927532
6250001	0.926179
6300001	0.926631
6350001	0.926366
6400001	0.925967
6450001	0.926117
6500001	0.925201
6550001	0.925491
6600001	0.925684
6650001	0.926242
6700001	0.925513
6750001	0.925049
6800001	0.924729
6850001	0.924436
6900001	0.92399
6950001	0.923515
7000001	0.923012
7050001	0.923181
7100001	0.922306
7150001	0.921971
7200001	0.922187
7250001	0.922155
7300001	0.922097
7350001	0.921756
7400001	0.922553
7450001	0.921305
7500001	0.921802
7550001	0.921665
7600001	0.920593
7650001	0.920876
7700001	0.920017
7750001	0.920445
7800001	0.920111
7850001	0.9198
7900001	0.919898
7950001	0.91879
8000001	0.920447
8050001	0.919781
8100001	0.919865
8150001	0.920796
8200001	0.920124
8250001	0.920438
8300001	0.920096
8350001	0.919398
8400001	0.919927
8450001	0.91883
8500001	0.918866
8550001	0.918194
8600001	0.918268
8650001	0.917588
8700001	0.918573
8750001	0.918102
8800001	0.917563
8850001	0.9187
8900001	0.918258
8950001	0.91771
9000001	0.917577
};
\addlegendentry{PA-I}

\addplot [mark=, densely dashdotted, line width=1.5pt, color=green] table {
1	2.50735
50001	1.57282
100001	1.36952
150001	1.26567
200001	1.20202
250001	1.15761
300001	1.12412
350001	1.09847
400001	1.07805
450001	1.06164
500001	1.04866
550001	1.03726
600001	1.02736
650001	1.01826
700001	1.01062
750001	1.0051
800001	0.998144
850001	0.993128
900001	0.987702
950001	0.984517
1000001	0.979482
1050001	0.975789
1100001	0.972537
1150001	0.970972
1200001	0.966507
1250001	0.964296
1300001	0.962587
1350001	0.960893
1400001	0.958182
1450001	0.95688
1500001	0.954086
1550001	0.952565
1600001	0.95028
1650001	0.94972
1700001	0.948301
1750001	0.946212
1800001	0.946287
1850001	0.945095
1900001	0.943844
1950001	0.941268
2000001	0.941925
2050001	0.939844
2100001	0.939402
2150001	0.93868
2200001	0.936859
2250001	0.935453
2300001	0.934362
2350001	0.934619
2400001	0.932365
2450001	0.932835
2500001	0.930892
2550001	0.931837
2600001	0.930412
2650001	0.930244
2700001	0.928528
2750001	0.92776
2800001	0.927945
2850001	0.927148
2900001	0.926064
2950001	0.925045
3000001	0.925403
3050001	0.92453
3100001	0.924576
3150001	0.922903
3200001	0.922799
3250001	0.922666
3300001	0.921711
3350001	0.923364
3400001	0.921111
3450001	0.919767
3500001	0.919104
3550001	0.919472
3600001	0.918316
3650001	0.918523
3700001	0.917445
3750001	0.917369
3800001	0.917234
3850001	0.916634
3900001	0.916377
3950001	0.918038
4000001	0.917142
4050001	0.915685
4100001	0.915604
4150001	0.914422
4200001	0.913365
4250001	0.912622
4300001	0.912477
4350001	0.911955
4400001	0.910875
4450001	0.912319
4500001	0.910359
4550001	0.909444
4600001	0.909669
4650001	0.90992
4700001	0.909892
4750001	0.909316
4800001	0.908935
4850001	0.906968
4900001	0.907477
4950001	0.908377
5000001	0.907877
5050001	0.906862
5100001	0.906519
5150001	0.905552
5200001	0.905958
5250001	0.905163
5300001	0.904183
5350001	0.904507
5400001	0.903636
5450001	0.904661
5500001	0.905737
5550001	0.904914
5600001	0.903735
5650001	0.902991
5700001	0.902583
5750001	0.901939
5800001	0.90179
5850001	0.901848
5900001	0.900698
5950001	0.901922
6000001	0.901435
6050001	0.900756
6100001	0.900267
6150001	0.900406
6200001	0.899411
6250001	0.89939
6300001	0.89875
6350001	0.899258
6400001	0.897591
6450001	0.897893
6500001	0.898282
6550001	0.897976
6600001	0.89782
6650001	0.897137
6700001	0.897518
6750001	0.896373
6800001	0.896279
6850001	0.895974
6900001	0.896038
6950001	0.894554
7000001	0.894348
7050001	0.895649
7100001	0.894047
7150001	0.89442
7200001	0.894106
7250001	0.8927
7300001	0.893375
7350001	0.894325
7400001	0.892956
7450001	0.89288
7500001	0.892912
7550001	0.893353
7600001	0.892059
7650001	0.893236
7700001	0.892191
7750001	0.892637
7800001	0.892009
7850001	0.892157
7900001	0.890269
7950001	0.890522
8000001	0.89082
8050001	0.891383
8100001	0.89184
8150001	0.891343
8200001	0.890571
8250001	0.890752
8300001	0.889642
8350001	0.890831
8400001	0.88863
8450001	0.889501
8500001	0.890257
8550001	0.888732
8600001	0.890215
8650001	0.889404
8700001	0.889316
8750001	0.888267
8800001	0.886948
8850001	0.888142
8900001	0.888568
8950001	0.888583
9000001	0.887386
};
\addlegendentry{CTR}
\addplot [mark=, ultra thick,line width=1.5pt, color=purple] table {
1	2.48686
50001	1.67167
100001	1.44304
150001	1.3271
200001	1.25621
250001	1.2049
300001	1.16645
350001	1.13693
400001	1.11258
450001	1.09256
500001	1.07689
550001	1.06314
600001	1.05069
650001	1.03964
700001	1.03006
750001	1.02244
800001	1.01489
850001	1.00834
900001	1.00195
950001	0.996805
1000001	0.991547
1050001	0.986398
1100001	0.982406
1150001	0.979457
1200001	0.974938
1250001	0.97177
1300001	0.969268
1350001	0.966661
1400001	0.963483
1450001	0.9615
1500001	0.958446
1550001	0.956737
1600001	0.953961
1650001	0.952808
1700001	0.951068
1750001	0.948779
1800001	0.947898
1850001	0.946452
1900001	0.945231
1950001	0.942845
2000001	0.942207
2050001	0.940256
2100001	0.939364
2150001	0.938362
2200001	0.936565
2250001	0.93559
2300001	0.934112
2350001	0.934191
2400001	0.932088
2450001	0.931773
2500001	0.930547
2550001	0.93089
2600001	0.929566
2650001	0.929108
2700001	0.927575
2750001	0.926783
2800001	0.926849
2850001	0.92597
2900001	0.925047
2950001	0.924225
3000001	0.924223
3050001	0.923897
3100001	0.923783
3150001	0.922418
3200001	0.922179
3250001	0.922012
3300001	0.921049
3350001	0.922152
3400001	0.920519
3450001	0.919693
3500001	0.918948
3550001	0.919252
3600001	0.918719
3650001	0.918213
3700001	0.917842
3750001	0.91782
3800001	0.917623
3850001	0.917304
3900001	0.917064
3950001	0.917875
4000001	0.917708
4050001	0.916513
4100001	0.916299
4150001	0.915495
4200001	0.915123
4250001	0.914652
4300001	0.914305
4350001	0.913826
4400001	0.913071
4450001	0.914106
4500001	0.913193
4550001	0.912153
4600001	0.912333
4650001	0.912709
4700001	0.912579
4750001	0.912267
4800001	0.912056
4850001	0.910947
4900001	0.911193
4950001	0.912054
5000001	0.911497
5050001	0.911536
5100001	0.910835
5150001	0.910344
5200001	0.910785
5250001	0.910104
5300001	0.909858
5350001	0.909874
5400001	0.909483
5450001	0.910208
5500001	0.910975
5550001	0.910677
5600001	0.909584
5650001	0.909396
5700001	0.909161
5750001	0.908973
5800001	0.908804
5850001	0.909003
5900001	0.908497
5950001	0.909062
6000001	0.908776
6050001	0.908749
6100001	0.908542
6150001	0.908342
6200001	0.908045
6250001	0.90772
6300001	0.90762
6350001	0.90775
6400001	0.906811
6450001	0.907293
6500001	0.907853
6550001	0.9076
6600001	0.90749
6650001	0.907137
6700001	0.907006
6750001	0.906601
6800001	0.906264
6850001	0.906293
6900001	0.906342
6950001	0.905828
7000001	0.905549
7050001	0.906526
7100001	0.905501
7150001	0.905493
7200001	0.905587
7250001	0.904795
7300001	0.905254
7350001	0.905394
7400001	0.904955
7450001	0.905199
7500001	0.905377
7550001	0.905597
7600001	0.904891
7650001	0.90546
7700001	0.90474
7750001	0.905447
7800001	0.904756
7850001	0.904969
7900001	0.903893
7950001	0.904525
8000001	0.904601
8050001	0.904932
8100001	0.905208
8150001	0.904997
8200001	0.904395
8250001	0.904717
8300001	0.90426
8350001	0.904969
8400001	0.903741
8450001	0.904086
8500001	0.904621
8550001	0.903649
8600001	0.904831
8650001	0.904098
8700001	0.904626
8750001	0.903718
8800001	0.902739
8850001	0.903832
8900001	0.904019
8950001	0.903732
9000001	0.903431
};
\addlegendentry{OCTR}
\addplot[mark=, ultra thick,line width=1.5pt, color=red] table {
1	2.55257
50001	1.2546
100001	1.14823
150001	1.09253
200001	1.05577
250001	1.0319
300001	1.01209
350001	0.99737
400001	0.985707
450001	0.976501
500001	0.96887
550001	0.962215
600001	0.956031
650001	0.950461
700001	0.945983
750001	0.942561
800001	0.93889
850001	0.935303
900001	0.932185
950001	0.929505
1000001	0.927027
1050001	0.924657
1100001	0.922356
1150001	0.920369
1200001	0.918803
1250001	0.917279
1300001	0.915904
1350001	0.91454
1400001	0.913149
1450001	0.911801
1500001	0.910341
1550001	0.909155
1600001	0.908024
1650001	0.90699
1700001	0.906275
1750001	0.905277
1800001	0.904396
1850001	0.903744
1900001	0.903185
1950001	0.902424
2000001	0.901499
2050001	0.900758
2100001	0.900177
2150001	0.899475
2200001	0.898743
2250001	0.898187
2300001	0.897571
2350001	0.897144
2400001	0.896563
2450001	0.896129
2500001	0.895604
2550001	0.895153
2600001	0.894664
2650001	0.894139
2700001	0.893622
2750001	0.893189
2800001	0.89281
2850001	0.892492
2900001	0.892021
2950001	0.891689
3000001	0.89141
3050001	0.891122
3100001	0.890846
3150001	0.890473
3200001	0.89008
3250001	0.88972
3300001	0.889404
3350001	0.889308
3400001	0.889021
3450001	0.888638
3500001	0.888313
3550001	0.888033
3600001	0.887829
3650001	0.887651
3700001	0.887426
3750001	0.887207
3800001	0.886943
3850001	0.886769
3900001	0.886667
3950001	0.886506
4000001	0.88636
4050001	0.886086
4100001	0.885834
4150001	0.885625
4200001	0.885376
4250001	0.885165
4300001	0.884901
4350001	0.884718
4400001	0.884584
4450001	0.88435
4500001	0.884218
4550001	0.883979
4600001	0.883823
4650001	0.883642
4700001	0.883502
4750001	0.883238
4800001	0.883139
4850001	0.882978
4900001	0.882899
4950001	0.88277
5000001	0.882593
5050001	0.882449
5100001	0.882293
5150001	0.882222
5200001	0.882128
5250001	0.881964
5300001	0.881869
5350001	0.881694
5400001	0.881574
5450001	0.881427
5500001	0.881275
5550001	0.881216
5600001	0.88109
5650001	0.880932
5700001	0.880765
5750001	0.88072
5800001	0.880673
5850001	0.880567
5900001	0.880519
5950001	0.880442
6000001	0.880372
6050001	0.880162
6100001	0.880065
6150001	0.88
6200001	0.879897
6250001	0.879738
6300001	0.879641
6350001	0.879581
6400001	0.879448
6450001	0.879369
6500001	0.879355
6550001	0.879278
6600001	0.879263
6650001	0.879213
6700001	0.879124
6750001	0.879035
6800001	0.879013
6850001	0.878966
6900001	0.878857
6950001	0.878818
7000001	0.87871
7050001	0.878643
7100001	0.878569
7150001	0.878506
7200001	0.878446
7250001	0.878339
7300001	0.878287
7350001	0.878262
7400001	0.878159
7450001	0.878056
7500001	0.877992
7550001	0.877944
7600001	0.877868
7650001	0.877797
7700001	0.877743
7750001	0.877707
7800001	0.877628
7850001	0.877583
7900001	0.877512
7950001	0.877439
8000001	0.877367
8050001	0.877286
8100001	0.877257
8150001	0.877218
8200001	0.877199
8250001	0.877158
8300001	0.877106
8350001	0.877023
8400001	0.87695
8450001	0.876878
8500001	0.876814
8550001	0.876736
8600001	0.876733
8650001	0.876647
8700001	0.876625
8750001	0.876571
8800001	0.876509
8850001	0.876471
8900001	0.876428
8950001	0.876417
9000001	0.876323
};
\addlegendentry{OBCTR}
\end{axis}
\end{tikzpicture} 

%% file: obctr_rmse_10.tex
\begin{tikzpicture}
\begin{axis}[
    xmode = log,        
    xmin=1e4, xmax=1e7,  
    xlabel = Number of Samples Seen,
    ylabel = RMSE,
    title = {$K=10$}
]
\addplot [mark=, densely dashdotted, line width=1.5pt, color=blue]table {
1	1.59551
50001	1.29436
100001	1.21403
150001	1.16614
200001	1.13288
250001	1.10815
300001	1.0882
350001	1.07178
400001	1.05904
450001	1.04787
500001	1.03851
550001	1.0296
600001	1.02189
650001	1.01374
700001	1.0083
750001	1.00312
800001	0.996733
850001	0.99284
900001	0.98924
950001	0.984945
1000001	0.982155
1050001	0.978055
1100001	0.974664
1150001	0.973701
1200001	0.969423
1250001	0.967282
1300001	0.965535
1350001	0.963109
1400001	0.960811
1450001	0.958428
1500001	0.956535
1550001	0.954161
1600001	0.953022
1650001	0.951995
1700001	0.950639
1750001	0.948698
1800001	0.946738
1850001	0.945655
1900001	0.94541
1950001	0.943392
2000001	0.943919
2050001	0.942131
2100001	0.940248
2150001	0.940221
2200001	0.938192
2250001	0.937617
2300001	0.936354
2350001	0.936186
2400001	0.934715
2450001	0.934839
2500001	0.933847
2550001	0.933129
2600001	0.931763
2650001	0.932446
2700001	0.930283
2750001	0.930742
2800001	0.930486
2850001	0.929361
2900001	0.929403
2950001	0.92728
3000001	0.927899
3050001	0.927468
3100001	0.927077
3150001	0.926205
3200001	0.926553
3250001	0.925115
3300001	0.925418
3350001	0.924472
3400001	0.925235
3450001	0.924028
3500001	0.9223
3550001	0.924235
3600001	0.923237
3650001	0.921476
3700001	0.922156
3750001	0.921992
3800001	0.921613
3850001	0.921627
3900001	0.922287
3950001	0.921175
4000001	0.92011
4050001	0.919829
4100001	0.919861
4150001	0.919479
4200001	0.919065
4250001	0.918631
4300001	0.91879
4350001	0.917276
4400001	0.918126
4450001	0.919882
4500001	0.91801
4550001	0.916797
4600001	0.915798
4650001	0.91761
4700001	0.916834
4750001	0.917731
4800001	0.916429
4850001	0.915408
4900001	0.9167
4950001	0.914995
5000001	0.915791
5050001	0.916114
5100001	0.914152
5150001	0.915478
5200001	0.914693
5250001	0.91469
5300001	0.914898
5350001	0.914872
5400001	0.914677
5450001	0.914633
5500001	0.915094
5550001	0.913642
5600001	0.914737
5650001	0.913956
5700001	0.913205
5750001	0.91378
5800001	0.91258
5850001	0.914428
5900001	0.913723
5950001	0.913159
6000001	0.913036
6050001	0.913558
6100001	0.913823
6150001	0.912506
6200001	0.912337
6250001	0.911459
6300001	0.911965
6350001	0.911692
6400001	0.912161
6450001	0.91194
6500001	0.910954
6550001	0.912112
6600001	0.911439
6650001	0.912162
6700001	0.911645
6750001	0.910915
6800001	0.91136
6850001	0.9111
6900001	0.910555
6950001	0.91017
7000001	0.910174
7050001	0.910591
7100001	0.909888
7150001	0.909006
7200001	0.910432
7250001	0.910231
7300001	0.910059
7350001	0.91039
7400001	0.910334
7450001	0.909093
7500001	0.910416
7550001	0.911033
7600001	0.908809
7650001	0.90985
7700001	0.908892
7750001	0.909697
7800001	0.909244
7850001	0.908743
7900001	0.908975
7950001	0.908079
8000001	0.909376
8050001	0.909092
8100001	0.908944
8150001	0.910807
8200001	0.909806
8250001	0.909598
8300001	0.909397
8350001	0.909473
8400001	0.909213
8450001	0.908634
8500001	0.909669
8550001	0.907687
8600001	0.90849
8650001	0.907929
8700001	0.909065
8750001	0.908263
8800001	0.907588
8850001	0.908956
8900001	0.908794
8950001	0.90874
9000001	0.908464

};
\addlegendentry{PA-I}

\addplot [mark=, densely dashdotted, line width=1.5pt, color=green]table {
1	1.48923
50001	1.21982
100001	1.14972
150001	1.10704
200001	1.07954
250001	1.05697
300001	1.03969
350001	1.02679
400001	1.01508
450001	1.00468
500001	0.997173
550001	0.990338
600001	0.983671
650001	0.977874
700001	0.972521
750001	0.968123
800001	0.964175
850001	0.960169
900001	0.957396
950001	0.95404
1000001	0.950514
1050001	0.947047
1100001	0.94547
1150001	0.944673
1200001	0.941545
1250001	0.939279
1300001	0.937968
1350001	0.936487
1400001	0.934162
1450001	0.933466
1500001	0.93134
1550001	0.930111
1600001	0.928651
1650001	0.92838
1700001	0.92702
1750001	0.925678
1800001	0.924729
1850001	0.924398
1900001	0.923508
1950001	0.921681
2000001	0.921798
2050001	0.920254
2100001	0.920414
2150001	0.919549
2200001	0.91819
2250001	0.917278
2300001	0.915847
2350001	0.916985
2400001	0.914933
2450001	0.91495
2500001	0.914345
2550001	0.915046
2600001	0.913282
2650001	0.913812
2700001	0.9124
2750001	0.911496
2800001	0.911833
2850001	0.910959
2900001	0.910339
2950001	0.909693
3000001	0.910143
3050001	0.909072
3100001	0.90965
3150001	0.908217
3200001	0.908504
3250001	0.907931
3300001	0.907495
3350001	0.908681
3400001	0.906504
3450001	0.905825
3500001	0.90528
3550001	0.905991
3600001	0.90558
3650001	0.905206
3700001	0.904957
3750001	0.904844
3800001	0.904465
3850001	0.904119
3900001	0.904272
3950001	0.905265
4000001	0.904606
4050001	0.903699
4100001	0.903744
4150001	0.903394
4200001	0.902738
4250001	0.902001
4300001	0.901585
4350001	0.901305
4400001	0.900518
4450001	0.901727
4500001	0.900773
4550001	0.899751
4600001	0.899708
4650001	0.900261
4700001	0.899951
4750001	0.8996
4800001	0.899144
4850001	0.898028
4900001	0.898556
4950001	0.899078
5000001	0.898797
5050001	0.898494
5100001	0.898277
5150001	0.897354
5200001	0.897609
5250001	0.897328
5300001	0.896367
5350001	0.896461
5400001	0.896071
5450001	0.896802
5500001	0.897734
5550001	0.896973
5600001	0.896377
5650001	0.895502
5700001	0.895455
5750001	0.895139
5800001	0.894761
5850001	0.89529
5900001	0.894197
5950001	0.895144
6000001	0.894623
6050001	0.894589
6100001	0.894415
6150001	0.89406
6200001	0.892941
6250001	0.893104
6300001	0.8932
6350001	0.892993
6400001	0.891984
6450001	0.892093
6500001	0.89265
6550001	0.892057
6600001	0.891749
6650001	0.891502
6700001	0.891406
6750001	0.890536
6800001	0.890442
6850001	0.889739
6900001	0.889791
6950001	0.889376
7000001	0.888728
7050001	0.889871
7100001	0.888325
7150001	0.888294
7200001	0.888914
7250001	0.887558
7300001	0.888092
7350001	0.888468
7400001	0.887397
7450001	0.887441
7500001	0.887551
7550001	0.887853
7600001	0.886385
7650001	0.886643
7700001	0.88588
7750001	0.88658
7800001	0.885978
7850001	0.885874
7900001	0.884708
7950001	0.885053
8000001	0.885339
8050001	0.885835
8100001	0.885689
8150001	0.885657
8200001	0.884722
8250001	0.88503
8300001	0.88413
8350001	0.884673
8400001	0.882979
8450001	0.8837
8500001	0.884395
8550001	0.883333
8600001	0.884064
8650001	0.883375
8700001	0.883449
8750001	0.882067
8800001	0.880993
8850001	0.882385
8900001	0.882388
8950001	0.881874
9000001	0.881187

};
\addlegendentry{CTR}

\addplot [mark=, ultra thick, line width=1.5pt,color=purple] table {
1	1.58322
50001	1.18174
100001	1.11315
150001	1.07484
200001	1.05014
250001	1.02886
300001	1.01512
350001	1.00412
400001	0.996569
450001	0.988752
500001	0.981915
550001	0.976501
600001	0.972382
650001	0.96835
700001	0.964262
750001	0.961361
800001	0.958413
850001	0.954876
900001	0.952874
950001	0.952091
1000001	0.948293
1050001	0.94643
1100001	0.945567
1150001	0.946072
1200001	0.943316
1250001	0.941742
1300001	0.940997
1350001	0.940821
1400001	0.938728
1450001	0.938322
1500001	0.936854
1550001	0.93672
1600001	0.934408
1650001	0.93559
1700001	0.93513
1750001	0.933426
1800001	0.934135
1850001	0.933611
1900001	0.932665
1950001	0.930661
2000001	0.932605
2050001	0.930613
2100001	0.931048
2150001	0.930861
2200001	0.928874
2250001	0.928057
2300001	0.927963
2350001	0.928965
2400001	0.92695
2450001	0.928379
2500001	0.926161
2550001	0.9275
2600001	0.926369
2650001	0.926882
2700001	0.925445
2750001	0.925379
2800001	0.925957
2850001	0.925042
2900001	0.924648
2950001	0.924117
3000001	0.925054
3050001	0.924618
3100001	0.924767
3150001	0.923407
3200001	0.923834
3250001	0.923968
3300001	0.923723
3350001	0.925649
3400001	0.922964
3450001	0.922444
3500001	0.921937
3550001	0.923416
3600001	0.922484
3650001	0.923161
3700001	0.922061
3750001	0.922526
3800001	0.9228
3850001	0.921834
3900001	0.922764
3950001	0.924933
4000001	0.923638
4050001	0.923079
4100001	0.92334
4150001	0.922832
4200001	0.921962
4250001	0.921355
4300001	0.92096
4350001	0.92012
4400001	0.919451
4450001	0.92146
4500001	0.919658
4550001	0.919509
4600001	0.919563
4650001	0.920619
4700001	0.921182
4750001	0.920331
4800001	0.920411
4850001	0.917966
4900001	0.919443
4950001	0.920763
5000001	0.920824
5050001	0.920216
5100001	0.920258
5150001	0.91849
5200001	0.92041
5250001	0.919609
5300001	0.917577
5350001	0.918439
5400001	0.917999
5450001	0.918998
5500001	0.921217
5550001	0.920002
5600001	0.919986
5650001	0.918349
5700001	0.918077
5750001	0.917957
5800001	0.917572
5850001	0.918188
5900001	0.916463
5950001	0.918668
6000001	0.917685
6050001	0.9179
6100001	0.917276
6150001	0.9181
6200001	0.916151
6250001	0.916525
6300001	0.916809
6350001	0.917604
6400001	0.915472
6450001	0.915654
6500001	0.916584
6550001	0.916671
6600001	0.916122
6650001	0.915697
6700001	0.916408
6750001	0.91481
6800001	0.914484
6850001	0.914179
6900001	0.914918
6950001	0.913449
7000001	0.912985
7050001	0.914704
7100001	0.912377
7150001	0.913507
7200001	0.913412
7250001	0.912434
7300001	0.912457
7350001	0.913315
7400001	0.912327
7450001	0.912189
7500001	0.911897
7550001	0.912275
7600001	0.910882
7650001	0.912028
7700001	0.911002
7750001	0.911625
7800001	0.911278
7850001	0.911333
7900001	0.909342
7950001	0.909763
8000001	0.909577
8050001	0.910652
8100001	0.910913
8150001	0.911143
8200001	0.909583
8250001	0.909787
8300001	0.908258
8350001	0.90931
8400001	0.907528
8450001	0.908833
8500001	0.909699
8550001	0.907449
8600001	0.909798
8650001	0.908489
8700001	0.908063
8750001	0.906805
8800001	0.905062
8850001	0.906548
8900001	0.907507
8950001	0.906586
9000001	0.905423

};
\addlegendentry{OCTR}

\addplot [mark=, ultra thick,line width=1.5pt, color=red]table {
1	1.60952
50001	1.18366
100001	1.10514
150001	1.06001
200001	1.03369
250001	1.01429
300001	0.999699
350001	0.989575
400001	0.979046
450001	0.971202
500001	0.965582
550001	0.958927
600001	0.954437
650001	0.95059
700001	0.946251
750001	0.942464
800001	0.940208
850001	0.937212
900001	0.93484
950001	0.932494
1000001	0.930569
1050001	0.928051
1100001	0.926419
1150001	0.924451
1200001	0.923196
1250001	0.921529
1300001	0.92013
1350001	0.91871
1400001	0.917512
1450001	0.916383
1500001	0.915386
1550001	0.914315
1600001	0.913309
1650001	0.912337
1700001	0.911418
1750001	0.910677
1800001	0.909681
1850001	0.908939
1900001	0.908054
1950001	0.907504
2000001	0.906929
2050001	0.906344
2100001	0.905765
2150001	0.905123
2200001	0.904457
2250001	0.903771
2300001	0.903295
2350001	0.902686
2400001	0.902333
2450001	0.90159
2500001	0.901119
2550001	0.900503
2600001	0.900023
2650001	0.899673
2700001	0.899292
2750001	0.898773
2800001	0.898105
2850001	0.897563
2900001	0.897243
2950001	0.896588
3000001	0.896164
3050001	0.89574
3100001	0.895423
3150001	0.895135
3200001	0.894973
3250001	0.894795
3300001	0.894476
3350001	0.894185
3400001	0.893881
3450001	0.893497
3500001	0.89305
3550001	0.892835
3600001	0.892449
3650001	0.892207
3700001	0.891984
3750001	0.891719
3800001	0.891448
3850001	0.891285
3900001	0.89114
3950001	0.890958
4000001	0.890664
4050001	0.89034
4100001	0.890145
4150001	0.889927
4200001	0.889823
4250001	0.88975
4300001	0.889478
4350001	0.889128
4400001	0.889004
4450001	0.888891
4500001	0.888684
4550001	0.888409
4600001	0.888149
4650001	0.88793
4700001	0.88779
4750001	0.887694
4800001	0.887636
4850001	0.887584
4900001	0.887249
4950001	0.886999
5000001	0.886797
5050001	0.886468
5100001	0.886363
5150001	0.886142
5200001	0.886114
5250001	0.886034
5300001	0.885925
5350001	0.885778
5400001	0.885684
5450001	0.885423
5500001	0.885191
5550001	0.885063
5600001	0.884972
5650001	0.884929
5700001	0.884755
5750001	0.884578
5800001	0.884579
5850001	0.884464
5900001	0.884325
5950001	0.884187
6000001	0.884035
6050001	0.883937
6100001	0.883851
6150001	0.883753
6200001	0.883774
6250001	0.883658
6300001	0.883508
6350001	0.883468
6400001	0.883298
6450001	0.883116
6500001	0.883017
6550001	0.882882
6600001	0.88282
6650001	0.882693
6700001	0.882744
6750001	0.88258
6800001	0.882483
6850001	0.882377
6900001	0.882249
6950001	0.882093
7000001	0.882022
7050001	0.881871
7100001	0.881828
7150001	0.881683
7200001	0.881509
7250001	0.881394
7300001	0.881295
7350001	0.881149
7400001	0.881113
7450001	0.88097
7500001	0.880758
7550001	0.880614
7600001	0.880592
7650001	0.880497
7700001	0.880434
7750001	0.880379
7800001	0.88029
7850001	0.880131
7900001	0.880149
7950001	0.880072
8000001	0.880061
8050001	0.880009
8100001	0.879912
8150001	0.879879
8200001	0.879873
8250001	0.879834
8300001	0.879761
8350001	0.879751
8400001	0.879693
8450001	0.879676
8500001	0.879637
8550001	0.879545
8600001	0.879511
8650001	0.879417
8700001	0.879364
8750001	0.879271
8800001	0.879126
8850001	0.878994
8900001	0.878945
8950001	0.878838
};
\addlegendentry{OBCTR}

\end{axis}
\end{tikzpicture} 

%% file: obctr_rmse_20.tex
\begin{tikzpicture}
\begin{axis}[
    xmode = log,        
    xmin=1e4, xmax=1e7,  
    xlabel = Number of Samples Seen,
    ylabel = RMSE,
    title = {$K=20$}
]
\addplot [mark=, densely dashdotted, line width=1.5pt, color=blue]table {
1	2.06559
50001	1.32768
100001	1.25535
150001	1.20825
200001	1.17527
250001	1.1494
300001	1.12907
350001	1.11369
400001	1.10033
450001	1.0861
500001	1.07646
550001	1.06618
600001	1.05766
650001	1.05114
700001	1.04497
750001	1.03668
800001	1.03201
850001	1.02633
900001	1.02138
950001	1.01637
1000001	1.01317
1050001	1.00838
1100001	1.00601
1150001	1.00229
1200001	0.999522
1250001	0.995906
1300001	0.99286
1350001	0.98986
1400001	0.987268
1450001	0.984644
1500001	0.982937
1550001	0.981131
1600001	0.978768
1650001	0.977784
1700001	0.974725
1750001	0.972606
1800001	0.971637
1850001	0.971134
1900001	0.96943
1950001	0.968523
2000001	0.968379
2050001	0.965377
2100001	0.963873
2150001	0.961573
2200001	0.960371
2250001	0.959783
2300001	0.957812
2350001	0.957085
2400001	0.956372
2450001	0.955476
2500001	0.953554
2550001	0.953332
2600001	0.952968
2650001	0.950617
2700001	0.950573
2750001	0.950472
2800001	0.948856
2850001	0.948433
2900001	0.946654
2950001	0.946359
3000001	0.945576
3050001	0.944786
3100001	0.944109
3150001	0.94495
3200001	0.943525
3250001	0.942951
3300001	0.941248
3350001	0.941607
3400001	0.940198
3450001	0.938615
3500001	0.941058
3550001	0.93965
3600001	0.937234
3650001	0.936784
3700001	0.938733
3750001	0.936361
3800001	0.936482
3850001	0.937165
3900001	0.937045
3950001	0.93517
4000001	0.934295
4050001	0.935058
4100001	0.935678
4150001	0.935035
4200001	0.932996
4250001	0.932274
4300001	0.930881
4350001	0.931224
4400001	0.93283
4450001	0.931343
4500001	0.930676
4550001	0.92983
4600001	0.931843
4650001	0.930851
4700001	0.93112
4750001	0.929522
4800001	0.928846
4850001	0.930439
4900001	0.928055
4950001	0.929577
5000001	0.928833
5050001	0.927625
5100001	0.927863
5150001	0.926502
5200001	0.925838
5250001	0.925793
5300001	0.925871
5350001	0.926407
5400001	0.926205
5450001	0.927823
5500001	0.925509
5550001	0.925682
5600001	0.92523
5650001	0.925192
5700001	0.925988
5750001	0.924062
5800001	0.925685
5850001	0.92364
5900001	0.923982
5950001	0.924008
6000001	0.923562
6050001	0.924786
6100001	0.923151
6150001	0.921895
6200001	0.921611
6250001	0.922291
6300001	0.922932
6350001	0.922711
6400001	0.922153
6450001	0.921261
6500001	0.923014
6550001	0.920822
6600001	0.920973
6650001	0.921232
6700001	0.920294
6750001	0.921341
6800001	0.920385
6850001	0.920078
6900001	0.919741
6950001	0.919329
7000001	0.919845
7050001	0.918924
7100001	0.918162
7150001	0.919476
7200001	0.919815
7250001	0.920304
7300001	0.919316
7350001	0.919107
7400001	0.917492
7450001	0.918783
7500001	0.919047
7550001	0.916372
7600001	0.91864
7650001	0.917831
7700001	0.918004
7750001	0.917921
7800001	0.91581
7850001	0.916845
7900001	0.915737
7950001	0.918047
8000001	0.917496
8050001	0.91626
8100001	0.919258
8150001	0.917905
8200001	0.918029
8250001	0.916073
8300001	0.916878
8350001	0.916849
8400001	0.915599
8450001	0.917384
8500001	0.914992
8550001	0.916082
8600001	0.914321
8650001	0.915994
8700001	0.914349
8750001	0.914604
8800001	0.916549
8850001	0.915341
8900001	0.916321
8950001	0.914767

};
\addlegendentry{PA-I}

\addplot [mark=, densely dashdotted, line width=1.5pt, color=green]table {
1	2.08584
50001	1.40143
100001	1.26646
150001	1.20202
200001	1.16425
250001	1.13598
300001	1.11404
350001	1.09829
400001	1.08385
450001	1.07046
500001	1.06189
550001	1.05227
600001	1.04387
650001	1.03644
700001	1.03026
750001	1.02293
800001	1.01781
850001	1.01226
900001	1.00771
950001	1.00325
1000001	0.998943
1050001	0.995589
1100001	0.991304
1150001	0.988134
1200001	0.985368
1250001	0.98196
1300001	0.979468
1350001	0.976262
1400001	0.97381
1450001	0.971739
1500001	0.968981
1550001	0.967243
1600001	0.966157
1650001	0.962514
1700001	0.960241
1750001	0.959308
1800001	0.95711
1850001	0.9558
1900001	0.953429
1950001	0.953353
2000001	0.950377
2050001	0.949818
2100001	0.948008
2150001	0.946616
2200001	0.946403
2250001	0.94369
2300001	0.943173
2350001	0.941627
2400001	0.940415
2450001	0.94031
2500001	0.939015
2550001	0.937033
2600001	0.93673
2650001	0.935751
2700001	0.934755
2750001	0.934158
2800001	0.933576
2850001	0.932706
2900001	0.931754
2950001	0.92991
3000001	0.929464
3050001	0.928625
3100001	0.927382
3150001	0.92697
3200001	0.926462
3250001	0.926043
3300001	0.925268
3350001	0.924309
3400001	0.923866
3450001	0.922483
3500001	0.922561
3550001	0.922904
3600001	0.922004
3650001	0.920911
3700001	0.920127
3750001	0.92004
3800001	0.919276
3850001	0.918082
3900001	0.919095
3950001	0.918296
4000001	0.917296
4050001	0.916847
4100001	0.917028
4150001	0.916661
4200001	0.915431
4250001	0.915595
4300001	0.915316
4350001	0.914258
4400001	0.913869
4450001	0.913595
4500001	0.91283
4550001	0.912547
4600001	0.912165
4650001	0.911383
4700001	0.911218
4750001	0.911164
4800001	0.910954
4850001	0.910083
4900001	0.910088
4950001	0.91006
5000001	0.910389
5050001	0.909088
5100001	0.909443
5150001	0.908668
5200001	0.908448
5250001	0.908436
5300001	0.907408
5350001	0.907929
5400001	0.907136
5450001	0.906719
5500001	0.906281
5550001	0.906458
5600001	0.905796
5650001	0.906509
5700001	0.905507
5750001	0.905596
5800001	0.905563
5850001	0.905313
5900001	0.905098
5950001	0.904058
6000001	0.903606
6050001	0.904623
6100001	0.903842
6150001	0.904043
6200001	0.903207
6250001	0.902245
6300001	0.903105
6350001	0.904296
6400001	0.90224
6450001	0.902047
6500001	0.901322
6550001	0.901702
6600001	0.902743
6650001	0.901755
6700001	0.900812
6750001	0.900855
6800001	0.901351
6850001	0.899914
6900001	0.900259
6950001	0.899389
7000001	0.900315
7050001	0.89998
7100001	0.900202
7150001	0.899611
7200001	0.898959
7250001	0.899341
7300001	0.899445
7350001	0.899306
7400001	0.898264
7450001	0.898284
7500001	0.898739
7550001	0.898236
7600001	0.898013
7650001	0.898012
7700001	0.897663
7750001	0.897394
7800001	0.897414
7850001	0.897574
7900001	0.897317
7950001	0.897816
8000001	0.897385
8050001	0.896544
8100001	0.897231
8150001	0.896716
8200001	0.897335
8250001	0.896415
8300001	0.896275
8350001	0.895824
8400001	0.895976
8450001	0.896099
8500001	0.895457
8550001	0.895608
8600001	0.896044
8650001	0.895001
8700001	0.894857
8750001	0.895563
8800001	0.89534
8850001	0.895179
8900001	0.895328
8950001	0.894675

};
\addlegendentry{CTR}

\addplot  [mark=, ultra thick, line width=1.5pt,color=purple] table {
1	2.06563
50001	1.15542
100001	1.11083
150001	1.08936
200001	1.06658
250001	1.04957
300001	1.0379
350001	1.0276
400001	1.01774
450001	1.01204
500001	1.00507
550001	0.998588
600001	0.99291
650001	0.991586
700001	0.987768
750001	0.986245
800001	0.978108
850001	0.976334
900001	0.975838
950001	0.968122
1000001	0.966903
1050001	0.966518
1100001	0.966672
1150001	0.962119
1200001	0.961297
1250001	0.959415
1300001	0.959219
1350001	0.955904
1400001	0.95592
1450001	0.953749
1500001	0.953455
1550001	0.950859
1600001	0.951367
1650001	0.951076
1700001	0.949384
1750001	0.950413
1800001	0.949287
1850001	0.945713
1900001	0.944638
1950001	0.946871
2000001	0.94489
2050001	0.944847
2100001	0.943485
2150001	0.942874
2200001	0.940256
2250001	0.940756
2300001	0.941154
2350001	0.938862
2400001	0.940447
2450001	0.937748
2500001	0.939212
2550001	0.936095
2600001	0.936887
2650001	0.937493
2700001	0.936014
2750001	0.935949
2800001	0.935471
2850001	0.934644
2900001	0.934338
2950001	0.934925
3000001	0.934582
3050001	0.934472
3100001	0.933005
3150001	0.932434
3200001	0.933209
3250001	0.933477
3300001	0.934829
3350001	0.931827
3400001	0.930935
3450001	0.92989
3500001	0.932777
3550001	0.930482
3600001	0.93256
3650001	0.930399
3700001	0.930577
3750001	0.931545
3800001	0.928814
3850001	0.930853
3900001	0.934197
3950001	0.931217
4000001	0.930334
4050001	0.930981
4100001	0.931783
4150001	0.929261
4200001	0.928756
4250001	0.928543
4300001	0.927526
4350001	0.926133
4400001	0.928961
4450001	0.926444
4500001	0.927619
4550001	0.926645
4600001	0.927632
4650001	0.928923
4700001	0.926954
4750001	0.927169
4800001	0.924222
4850001	0.925684
4900001	0.926861
4950001	0.92795
5000001	0.926071
5050001	0.926845
5100001	0.924386
5150001	0.926994
5200001	0.925778
5250001	0.922254
5300001	0.924295
5350001	0.92378
5400001	0.924149
5450001	0.92729
5500001	0.925057
5550001	0.927162
5600001	0.923895
5650001	0.923957
5700001	0.923581
5750001	0.922916
5800001	0.923611
5850001	0.921458
5900001	0.924117
5950001	0.923349
6000001	0.922546
6050001	0.92234
6100001	0.923542
6150001	0.919586
6200001	0.920246
6250001	0.921656
6300001	0.923361
6350001	0.920073
6400001	0.919362
6450001	0.920989
6500001	0.921175
6550001	0.920592
6600001	0.920528
6650001	0.92198
6700001	0.918798
6750001	0.919358
6800001	0.917687
6850001	0.918701
6900001	0.917188
6950001	0.916436
7000001	0.918571
7050001	0.915681
7100001	0.917605
7150001	0.917769
7200001	0.9158
7250001	0.915407
7300001	0.91768
7350001	0.916755
7400001	0.916092
7450001	0.914828
7500001	0.916029
7550001	0.913081
7600001	0.914214
7650001	0.914053
7700001	0.914438
7750001	0.914399
7800001	0.914463
7850001	0.911842
7900001	0.912386
7950001	0.911992
8000001	0.91472
8050001	0.913515
8100001	0.914886
8150001	0.912632
8200001	0.912861
8250001	0.910849
8300001	0.912162
8350001	0.909867
8400001	0.912633
8450001	0.911355
8500001	0.909579
8550001	0.913351
8600001	0.912275
8650001	0.910852
8700001	0.908933
8750001	0.906893
8800001	0.908978
8850001	0.910379
8900001	0.909242
8950001	0.908506

};
\addlegendentry{OCTR}

\addplot [mark=, ultra thick, line width=1.5pt,color=red] table {
1	2.07991
50001	1.18215
100001	1.10827
150001	1.06961
200001	1.04692
250001	1.0296
300001	1.01743
350001	1.00843
400001	0.998177
450001	0.990915
500001	0.985729
550001	0.979555
600001	0.975036
650001	0.971181
700001	0.967155
750001	0.963708
800001	0.961534
850001	0.957785
900001	0.955363
950001	0.95233
1000001	0.950446
1050001	0.94748
1100001	0.94561
1150001	0.942999
1200001	0.941272
1250001	0.939414
1300001	0.937653
1350001	0.935743
1400001	0.93406
1450001	0.932767
1500001	0.931752
1550001	0.930348
1600001	0.92902
1650001	0.927921
1700001	0.926784
1750001	0.925791
1800001	0.924164
1850001	0.923334
1900001	0.922073
1950001	0.921334
2000001	0.920532
2050001	0.919524
2100001	0.918816
2150001	0.917922
2200001	0.916969
2250001	0.916016
2300001	0.915502
2350001	0.914402
2400001	0.913849
2450001	0.912823
2500001	0.912193
2550001	0.911225
2600001	0.910501
2650001	0.910092
2700001	0.909521
2750001	0.908762
2800001	0.907923
2850001	0.907042
2900001	0.906462
2950001	0.905558
3000001	0.904791
3050001	0.904175
3100001	0.903728
3150001	0.903356
3200001	0.903128
3250001	0.902921
3300001	0.902333
3350001	0.901884
3400001	0.901398
3450001	0.900748
3500001	0.900104
3550001	0.899726
3600001	0.89905
3650001	0.898713
3700001	0.898389
3750001	0.897992
3800001	0.897553
3850001	0.897122
3900001	0.896904
3950001	0.896528
4000001	0.895954
4050001	0.895435
4100001	0.895168
4150001	0.894791
4200001	0.894556
4250001	0.89442
4300001	0.893919
4350001	0.893324
4400001	0.893071
4450001	0.892964
4500001	0.892535
4550001	0.892135
4600001	0.891646
4650001	0.891315
4700001	0.89104
4750001	0.890937
4800001	0.890912
4850001	0.890771
4900001	0.890258
4950001	0.889825
5000001	0.889467
5050001	0.889001
5100001	0.888774
5150001	0.88844
5200001	0.888408
5250001	0.888288
5300001	0.888012
5350001	0.887761
5400001	0.887517
5450001	0.887098
5500001	0.886628
5550001	0.886388
5600001	0.886179
5650001	0.886077
5700001	0.885756
5750001	0.885399
5800001	0.885403
5850001	0.885223
5900001	0.884995
5950001	0.8848
6000001	0.884571
6050001	0.88442
6100001	0.884235
6150001	0.884092
6200001	0.884056
6250001	0.88385
6300001	0.883567
6350001	0.883493
6400001	0.883196
6450001	0.882858
6500001	0.882727
6550001	0.882455
6600001	0.882326
6650001	0.882073
6700001	0.88216
6750001	0.881769
6800001	0.881607
6850001	0.881428
6900001	0.881169
6950001	0.880917
7000001	0.880779
7050001	0.880546
7100001	0.88045
7150001	0.880143
7200001	0.879856
7250001	0.879663
7300001	0.87949
7350001	0.879236
7400001	0.879149
7450001	0.878958
7500001	0.878572
7550001	0.878377
7600001	0.878317
7650001	0.878145
7700001	0.877996
7750001	0.877878
7800001	0.87771
7850001	0.87744
7900001	0.877445
7950001	0.87728
8000001	0.877206
8050001	0.87708
8100001	0.876856
8150001	0.876763
8200001	0.876742
8250001	0.876666
8300001	0.876536
8350001	0.876461
8400001	0.876307
8450001	0.876225
8500001	0.876127
8550001	0.875957
8600001	0.875877
8650001	0.875692
8700001	0.875552
8750001	0.875411
8800001	0.875179
8850001	0.874969
8900001	0.874877
8950001	0.874708
};
\addlegendentry{OBCTR}

\end{axis}
\end{tikzpicture}

%% file: likelihood.tex
\begin{tikzpicture}
\begin{axis}[
    xmode = log,        
    xmin=1e4, xmax=1e7,  
    xlabel = Number of Samples Seen,
    ylabel = \small{Held-out Log-likelihood},
    legend style={legend pos=north west,font=\tiny},
    title = {}
]
\addplot[color=violet, mark=, densely dashdotted, line width=1.5pt,] table{
1	-8.756730833
50001	-6.297949167
100001	-6.256140833
150001	-6.229001667
200001	-6.20996
250001	-6.196240833
300001	-6.189226667
350001	-6.1839775
400001	-6.178503333
450001	-6.174778333
500001	-6.170435833
550001	-6.1659025
600001	-6.16329
650001	-6.160239167
700001	-6.156559167
750001	-6.155105833
800001	-6.152711667
850001	-6.150849167
900001	-6.1491825
950001	-6.146705833
1000001	-6.145438333
1050001	-6.144536667
1100001	-6.143360833
1150001	-6.142218333
1200001	-6.141656667
1250001	-6.1402575
1300001	-6.139614167
1350001	-6.1386675
1400001	-6.1381875
1450001	-6.137659167
1500001	-6.136388333
1550001	-6.1359275
1600001	-6.135535833
1650001	-6.133460833
1700001	-6.132365833
1750001	-6.131816667
1800001	-6.131556667
1850001	-6.131083333
1900001	-6.130665
1950001	-6.13012
2000001	-6.129599167
2050001	-6.129285
2100001	-6.1283975
2150001	-6.128346667
2200001	-6.12804
2250001	-6.127616667
2300001	-6.12738
2350001	-6.1270825
2400001	-6.12672
2450001	-6.126728333
2500001	-6.126546667
2550001	-6.1261025
2600001	-6.125830833
2650001	-6.125160833
2700001	-6.124878333
2750001	-6.124675
2800001	-6.124436667
2850001	-6.1239925
2900001	-6.123994167
2950001	-6.123355833
3000001	-6.123043333
3050001	-6.122899167
3100001	-6.122678333
3150001	-6.122399167
3200001	-6.1221525
3250001	-6.121988333
3300001	-6.121929167
3350001	-6.121616667
3400001	-6.121669167
3450001	-6.121145
3500001	-6.121359167
3550001	-6.120984167
3600001	-6.120669167
3650001	-6.120645
3700001	-6.120540833
3750001	-6.120256667
3800001	-6.120226667
3850001	-6.120125833
3900001	-6.120184167
3950001	-6.119985
4000001	-6.119848333
4050001	-6.119450833
4100001	-6.1194275
4150001	-6.119259167
4200001	-6.11908
4250001	-6.119005833
4300001	-6.119061667
4350001	-6.118659167
4400001	-6.118635833
4450001	-6.118683333
4500001	-6.118514167
4550001	-6.118566667
4600001	-6.11857
4650001	-6.118140833
4700001	-6.1183275
4750001	-6.118175
4800001	-6.117994167
4850001	-6.117899167
4900001	-6.117745
4950001	-6.1176425
5000001	-6.117531667
5050001	-6.117349167
5100001	-6.117435833
5150001	-6.117190833
5200001	-6.117185
5250001	-6.117096667
5300001	-6.117129167
5350001	-6.116768333
5400001	-6.116756667
5450001	-6.116583333
5500001	-6.1165525
5550001	-6.116543333
5600001	-6.116436667
5650001	-6.116394167
5700001	-6.116225
5750001	-6.1161975
5800001	-6.116014167
5850001	-6.115875
5900001	-6.1160575
5950001	-6.115775833
6000001	-6.1157275
6050001	-6.1158425
6100001	-6.115808333
6150001	-6.115795
6200001	-6.115709167
6250001	-6.115570833
6300001	-6.115455833
6350001	-6.115363333
6400001	-6.1151275
6450001	-6.1152925
6500001	-6.115058333
6550001	-6.1150625
6600001	-6.114691667
6650001	-6.114858333
6700001	-6.114846667
6750001	-6.114656667
6800001	-6.114774167
6850001	-6.114625
6900001	-6.1144375
6950001	-6.114421667
7000001	-6.114414167
7050001	-6.114355
7100001	-6.114348333
7150001	-6.1143675
7200001	-6.114239167
7250001	-6.1140375
7300001	-6.113976667
7350001	-6.114103333
7400001	-6.1138925
7450001	-6.1138975
7500001	-6.113684167
7550001	-6.113730833
7600001	-6.113865
7650001	-6.113674167
7700001	-6.11375
7750001	-6.113463333
7800001	-6.113405833
7850001	-6.1133275
7900001	-6.1132525
7950001	-6.1133225
8000001	-6.113168333
8050001	-6.1130025
8100001	-6.113193333
8150001	-6.113008333
8200001	-6.112984167
8250001	-6.112846667
8300001	-6.112899167
8350001	-6.112989167
8400001	-6.112995
8450001	-6.112785833
8500001	-6.112978333
8550001	-6.11273
8600001	-6.112658333
8650001	-6.11252
8700001	-6.1126875
8750001	-6.112488333
8800001	-6.112341667
8850001	-6.112425833
8900001	-6.112535833
8950001	-6.112246667

};
\addlegendentry{K=5, Online-LDA}
\addplot [color=red, mark=, ultra thick] table {1	-8.6782
50001	-6.20387
100001	-6.1689
150001	-6.14589
200001	-6.12734
250001	-6.11507
300001	-6.10999
350001	-6.10725
400001	-6.10299
450001	-6.10143
500001	-6.09849
550001	-6.0952
600001	-6.09406
650001	-6.09153
700001	-6.08686
750001	-6.08629
800001	-6.08424
850001	-6.0827
900001	-6.08107
950001	-6.07857
1000001	-6.07845
1050001	-6.07843
1100001	-6.07673
1150001	-6.07593
1200001	-6.07578
1250001	-6.07537
1300001	-6.07482
1350001	-6.07475
1400001	-6.07423
1450001	-6.07402
1500001	-6.07248
1550001	-6.07271
1600001	-6.07243
1650001	-6.07144
1700001	-6.07014
1750001	-6.06955
1800001	-6.06975
1850001	-6.06877
1900001	-6.06936
1950001	-6.06942
2000001	-6.06837
2050001	-6.06878
2100001	-6.06864
2150001	-6.0683
2200001	-6.06818
2250001	-6.06739
2300001	-6.06825
2350001	-6.06763
2400001	-6.06808
2450001	-6.06824
2500001	-6.06774
2550001	-6.06777
2600001	-6.06765
2650001	-6.06754
2700001	-6.06697
2750001	-6.06675
2800001	-6.06655
2850001	-6.06682
2900001	-6.06637
2950001	-6.06594
3000001	-6.06531
3050001	-6.06568
3100001	-6.06589
3150001	-6.06546
3200001	-6.06532
3250001	-6.06507
3300001	-6.06523
3350001	-6.06484
3400001	-6.0653
3450001	-6.06552
3500001	-6.06477
3550001	-6.065
3600001	-6.06475
3650001	-6.06487
3700001	-6.06496
3750001	-6.06438
3800001	-6.06387
3850001	-6.06457
3900001	-6.06433
3950001	-6.06395
4000001	-6.06415
4050001	-6.06446
4100001	-6.06434
4150001	-6.06403
4200001	-6.06388
4250001	-6.06398
4300001	-6.06394
4350001	-6.06407
4400001	-6.06385
4450001	-6.0641
4500001	-6.06357
4550001	-6.06397
4600001	-6.06399
4650001	-6.06402
4700001	-6.06397
4750001	-6.06351
4800001	-6.06388
4850001	-6.06352
4900001	-6.06388
4950001	-6.06439
5000001	-6.06391
5050001	-6.06419
5100001	-6.06396
5150001	-6.0637
5200001	-6.06427
5250001	-6.06422
5300001	-6.06367
5350001	-6.0644
5400001	-6.0634
5450001	-6.06403
5500001	-6.06327
5550001	-6.06361
5600001	-6.06338
5650001	-6.06357
5700001	-6.06341
5750001	-6.06329
5800001	-6.0636
5850001	-6.06338
5900001	-6.06417
5950001	-6.06375
6000001	-6.06367
6050001	-6.06365
6100001	-6.06334
6150001	-6.06385
6200001	-6.06361
6250001	-6.06346
6300001	-6.06328
6350001	-6.06364
6400001	-6.06295
6450001	-6.06309
6500001	-6.06325
6550001	-6.06284
6600001	-6.063
6650001	-6.06276
6700001	-6.06324
6750001	-6.06267
6800001	-6.06287
6850001	-6.06218
6900001	-6.06273
6950001	-6.0632
7000001	-6.06306
7050001	-6.0633
7100001	-6.06362
7150001	-6.06352
7200001	-6.06311
7250001	-6.06332
7300001	-6.06321
7350001	-6.06299
7400001	-6.06265
7450001	-6.06258
7500001	-6.06276
7550001	-6.06301
7600001	-6.06373
7650001	-6.06327
7700001	-6.06374
7750001	-6.06322
7800001	-6.06287
7850001	-6.0634
7900001	-6.06319
7950001	-6.06279
8000001	-6.06297
8050001	-6.06324
8100001	-6.0631
8150001	-6.06322
8200001	-6.06301
8250001	-6.06381
8300001	-6.0633
8350001	-6.06312
8400001	-6.06271
8450001	-6.06308
8500001	-6.06352
8550001	-6.06381
8600001	-6.06367
8650001	-6.06299
8700001	-6.06362
8750001	-6.06294
8800001	-6.06294
8850001	-6.06339
8900001	-6.06325
8950001	-6.06347
};
\addlegendentry{K=5, OBCTR}

\addplot[color=blue, mark=, densely dashdotted, line width=1.5pt,] table {
1	-8.815703636
50001	-6.279448182
100001	-6.207363636
150001	-6.1613
200001	-6.128923636
250001	-6.105561818
300001	-6.093007273
350001	-6.08325
400001	-6.074201818
450001	-6.067274545
500001	-6.060573636
550001	-6.052218182
600001	-6.047944545
650001	-6.042595455
700001	-6.036800909
750001	-6.034201818
800001	-6.029250909
850001	-6.026620909
900001	-6.023568182
950001	-6.019363636
1000001	-6.017208182
1050001	-6.015865455
1100001	-6.013566364
1150001	-6.012125455
1200001	-6.01061
1250001	-6.007881818
1300001	-6.007058182
1350001	-6.005401818
1400001	-6.00462
1450001	-6.00363
1500001	-6.00185
1550001	-6.001204545
1600001	-6.000042727
1650001	-5.997262727
1700001	-5.995953636
1750001	-5.994522727
1800001	-5.99422
1850001	-5.993464545
1900001	-5.992909091
1950001	-5.99224
2000001	-5.99135
2050001	-5.990854545
2100001	-5.989118182
2150001	-5.988739091
2200001	-5.988028182
2250001	-5.987759091
2300001	-5.987450909
2350001	-5.987
2400001	-5.986703636
2450001	-5.986188182
2500001	-5.985652727
2550001	-5.984964545
2600001	-5.984639091
2650001	-5.983845455
2700001	-5.983417273
2750001	-5.982900909
2800001	-5.982609091
2850001	-5.982355455
2900001	-5.981986364
2950001	-5.980906364
3000001	-5.980230909
3050001	-5.979789091
3100001	-5.979745455
3150001	-5.978936364
3200001	-5.97851
3250001	-5.978408182
3300001	-5.978263636
3350001	-5.977908182
3400001	-5.977527273
3450001	-5.97723
3500001	-5.977200909
3550001	-5.976879091
3600001	-5.976388182
3650001	-5.976226364
3700001	-5.975686364
3750001	-5.975260909
3800001	-5.975145455
3850001	-5.975116364
3900001	-5.974845455
3950001	-5.974891818
4000001	-5.974351818
4050001	-5.974304545
4100001	-5.973903636
4150001	-5.974098182
4200001	-5.973580909
4250001	-5.973473636
4300001	-5.972973636
4350001	-5.972816364
4400001	-5.972768182
4450001	-5.972544545
4500001	-5.972362727
4550001	-5.972269091
4600001	-5.972061818
4650001	-5.971777273
4700001	-5.971416364
4750001	-5.971269091
4800001	-5.971144545
4850001	-5.970951818
4900001	-5.970946364
4950001	-5.970774545
5000001	-5.970105455
5050001	-5.970411818
5100001	-5.970361818
5150001	-5.97021
5200001	-5.969854545
5250001	-5.969662727
5300001	-5.969857273
5350001	-5.969552727
5400001	-5.969312727
5450001	-5.969029091
5500001	-5.969063636
5550001	-5.968847273
5600001	-5.968591818
5650001	-5.968545455
5700001	-5.968555455
5750001	-5.968293636
5800001	-5.967806364
5850001	-5.968077273
5900001	-5.967955455
5950001	-5.967557273
6000001	-5.967489091
6050001	-5.967508182
6100001	-5.967554545
6150001	-5.967196364
6200001	-5.96682
6250001	-5.966891818
6300001	-5.966732727
6350001	-5.967029091
6400001	-5.966809091
6450001	-5.966470909
6500001	-5.966487273
6550001	-5.966159091
6600001	-5.966172727
6650001	-5.966044545
6700001	-5.965976364
6750001	-5.965768182
6800001	-5.965520909
6850001	-5.965596364
6900001	-5.965511818
6950001	-5.965516364
7000001	-5.965471818
7050001	-5.965343636
7100001	-5.965327273
7150001	-5.965004545
7200001	-5.964727273
7250001	-5.964848182
7300001	-5.964457273
7350001	-5.964466364
7400001	-5.964341818
7450001	-5.964092727
7500001	-5.964072727
7550001	-5.964041818
7600001	-5.963985455
7650001	-5.96368
7700001	-5.963420909
7750001	-5.963565455
7800001	-5.963556364
7850001	-5.963484545
7900001	-5.963502727
7950001	-5.963500909
8000001	-5.963430909
8050001	-5.963253636
8100001	-5.962964545
8150001	-5.96311
8200001	-5.962735455
8250001	-5.962823636
8300001	-5.962805455
8350001	-5.962661818
8400001	-5.962546364
8450001	-5.962672727
8500001	-5.962648182
8550001	-5.962480909
8600001	-5.962332727
8650001	-5.962372727
8700001	-5.962194545
8750001	-5.962116364
8800001	-5.962099091
8850001	-5.962098182
8900001	-5.961826364
8950001	-5.96179

};
\addlegendentry{K=10, Online-LDA}
\addplot [color=cyan, mark=, ultra thick] table {
1	-8.68057
50001	-6.13274
100001	-6.07627
150001	-6.0419
200001	-6.01438
250001	-5.99691
300001	-5.98851
350001	-5.98222
400001	-5.97488
450001	-5.96972
500001	-5.96477
550001	-5.9565
600001	-5.95418
650001	-5.95036
700001	-5.94556
750001	-5.94363
800001	-5.93935
850001	-5.93801
900001	-5.93708
950001	-5.93188
1000001	-5.92984
1050001	-5.9293
1100001	-5.92925
1150001	-5.92725
1200001	-5.92617
1250001	-5.92356
1300001	-5.92384
1350001	-5.92283
1400001	-5.92281
1450001	-5.92181
1500001	-5.92103
1550001	-5.91953
1600001	-5.91948
1650001	-5.91661
1700001	-5.91598
1750001	-5.91542
1800001	-5.91537
1850001	-5.9149
1900001	-5.9144
1950001	-5.91374
2000001	-5.91315
2050001	-5.91367
2100001	-5.91087
2150001	-5.91029
2200001	-5.9102
2250001	-5.90992
2300001	-5.91034
2350001	-5.91001
2400001	-5.91019
2450001	-5.90916
2500001	-5.90969
2550001	-5.90922
2600001	-5.90959
2650001	-5.90834
2700001	-5.90895
2750001	-5.90857
2800001	-5.90734
2850001	-5.90746
2900001	-5.90744
2950001	-5.90633
3000001	-5.90552
3050001	-5.90514
3100001	-5.90459
3150001	-5.90379
3200001	-5.90412
3250001	-5.90355
3300001	-5.90378
3350001	-5.90324
3400001	-5.90381
3450001	-5.90337
3500001	-5.9035
3550001	-5.90335
3600001	-5.90308
3650001	-5.9029
3700001	-5.90243
3750001	-5.90251
3800001	-5.90203
3850001	-5.90251
3900001	-5.90241
3950001	-5.902
4000001	-5.902
4050001	-5.90294
4100001	-5.90234
4150001	-5.90233
4200001	-5.90244
4250001	-5.90157
4300001	-5.90155
4350001	-5.90171
4400001	-5.90209
4450001	-5.90225
4500001	-5.90183
4550001	-5.9021
4600001	-5.90165
4650001	-5.90166
4700001	-5.90131
4750001	-5.90235
4800001	-5.90154
4850001	-5.90112
4900001	-5.90169
4950001	-5.90134
5000001	-5.90076
5050001	-5.90046
5100001	-5.9007
5150001	-5.90021
5200001	-5.90077
5250001	-5.9004
5300001	-5.9
5350001	-5.89943
5400001	-5.90008
5450001	-5.90013
5500001	-5.89952
5550001	-5.8998
5600001	-5.89916
5650001	-5.89936
5700001	-5.89943
5750001	-5.89931
5800001	-5.89875
5850001	-5.89887
5900001	-5.89921
5950001	-5.89915
6000001	-5.89958
6050001	-5.89954
6100001	-5.89863
6150001	-5.89899
6200001	-5.89888
6250001	-5.8981
6300001	-5.89765
6350001	-5.89782
6400001	-5.89802
6450001	-5.89841
6500001	-5.89795
6550001	-5.8981
6600001	-5.89875
6650001	-5.89757
6700001	-5.89794
6750001	-5.89814
6800001	-5.89789
6850001	-5.89752
6900001	-5.8979
6950001	-5.89801
7000001	-5.89832
7050001	-5.89788
7100001	-5.89791
7150001	-5.89761
7200001	-5.89769
7250001	-5.89732
7300001	-5.89719
7350001	-5.89745
7400001	-5.89659
7450001	-5.89764
7500001	-5.89674
7550001	-5.89698
7600001	-5.89662
7650001	-5.89666
7700001	-5.89656
7750001	-5.89812
7800001	-5.89687
7850001	-5.8967
7900001	-5.89742
7950001	-5.89689
8000001	-5.89677
8050001	-5.89702
8100001	-5.89694
8150001	-5.89701
8200001	-5.89732
8250001	-5.89778
8300001	-5.89698
8350001	-5.89652
8400001	-5.89736
8450001	-5.89656
8500001	-5.8972
8550001	-5.896
8600001	-5.89649
8650001	-5.89669
8700001	-5.89657
8750001	-5.89619
8800001	-5.89637
8850001	-5.89658
8900001	-5.89652
8950001	-5.89648

};
\addlegendentry{K=10, OBCTR}

\addplot [color=green, mark=, densely dashdotted, line width=1.5pt,] table {
1	-8.863862222
50001	-6.215218889
100001	-6.113386667
150001	-6.052751111
200001	-6.012855556
250001	-5.984425556
300001	-5.969221111
350001	-5.95688
400001	-5.945827778
450001	-5.937478889
500001	-5.929432222
550001	-5.920293333
600001	-5.914817778
650001	-5.908493333
700001	-5.901707778
750001	-5.898682222
800001	-5.892983333
850001	-5.889423333
900001	-5.885786667
950001	-5.880458889
1000001	-5.878175556
1050001	-5.876651111
1100001	-5.874263333
1150001	-5.872457778
1200001	-5.870701111
1250001	-5.868113333
1300001	-5.866962222
1350001	-5.864967778
1400001	-5.863943333
1450001	-5.863072222
1500001	-5.860786667
1550001	-5.860462222
1600001	-5.859205556
1650001	-5.855442222
1700001	-5.854066667
1750001	-5.852262222
1800001	-5.851617778
1850001	-5.851125556
1900001	-5.849863333
1950001	-5.849277778
2000001	-5.84818
2050001	-5.847606667
2100001	-5.845848889
2150001	-5.845594444
2200001	-5.845067778
2250001	-5.844428889
2300001	-5.843913333
2350001	-5.84361
2400001	-5.843276667
2450001	-5.842518889
2500001	-5.841973333
2550001	-5.841561111
2600001	-5.841332222
2650001	-5.840257778
2700001	-5.839744444
2750001	-5.839374444
2800001	-5.838956667
2850001	-5.838018889
2900001	-5.837578889
2950001	-5.836278889
3000001	-5.835985556
3050001	-5.835091111
3100001	-5.835247778
3150001	-5.834754444
3200001	-5.834168889
3250001	-5.83372
3300001	-5.833554444
3350001	-5.83346
3400001	-5.832941111
3450001	-5.832876667
3500001	-5.832757778
3550001	-5.832483333
3600001	-5.831682222
3650001	-5.831487778
3700001	-5.831066667
3750001	-5.830706667
3800001	-5.830461111
3850001	-5.829946667
3900001	-5.829952222
3950001	-5.829807778
4000001	-5.829817778
4050001	-5.8294
4100001	-5.829098889
4150001	-5.828894444
4200001	-5.829203333
4250001	-5.82877
4300001	-5.828464444
4350001	-5.828267778
4400001	-5.828342222
4450001	-5.828178889
4500001	-5.827852222
4550001	-5.827605556
4600001	-5.82722
4650001	-5.826772222
4700001	-5.827144444
4750001	-5.826672222
4800001	-5.826555556
4850001	-5.826485556
4900001	-5.826332222
4950001	-5.826553333
5000001	-5.825946667
5050001	-5.825764444
5100001	-5.826098889
5150001	-5.825665556
5200001	-5.825553333
5250001	-5.82507
5300001	-5.825071111
5350001	-5.824854444
5400001	-5.824292222
5450001	-5.824337778
5500001	-5.823962222
5550001	-5.823883333
5600001	-5.82383
5650001	-5.823806667
5700001	-5.823506667
5750001	-5.823155556
5800001	-5.822993333
5850001	-5.822847778
5900001	-5.822907778
5950001	-5.822574444
6000001	-5.82285
6050001	-5.822752222
6100001	-5.822623333
6150001	-5.822368889
6200001	-5.82239
6250001	-5.822038889
6300001	-5.822122222
6350001	-5.822115556
6400001	-5.82155
6450001	-5.821422222
6500001	-5.821374444
6550001	-5.821494444
6600001	-5.821218889
6650001	-5.820932222
6700001	-5.821292222
6750001	-5.820824444
6800001	-5.820703333
6850001	-5.820667778
6900001	-5.820795556
6950001	-5.820772222
7000001	-5.820245556
7050001	-5.820601111
7100001	-5.820501111
7150001	-5.819925556
7200001	-5.82014
7250001	-5.819995556
7300001	-5.819917778
7350001	-5.819972222
7400001	-5.819718889
7450001	-5.819712222
7500001	-5.819622222
7550001	-5.81982
7600001	-5.819588889
7650001	-5.81943
7700001	-5.819227778
7750001	-5.818942222
7800001	-5.819021111
7850001	-5.818882222
7900001	-5.819093333
7950001	-5.818384444
8000001	-5.818585556
8050001	-5.818546667
8100001	-5.818282222
8150001	-5.818393333
8200001	-5.818546667
8250001	-5.818227778
8300001	-5.817814444
8350001	-5.817878889
8400001	-5.817524444
8450001	-5.818057778
8500001	-5.817802222
8550001	-5.818097778
8600001	-5.817672222
8650001	-5.817371111
8700001	-5.81743
8750001	-5.817472222
8800001	-5.817521111
8850001	-5.817266667
8900001	-5.816745556
8950001	-5.816968889

};
\addlegendentry{K=20, Online-LDA}
\addplot [color=teal, mark=, ultra thick] table {
1	-8.68176
50001	-6.0301
100001	-5.96831
150001	-5.91208
200001	-5.88099
250001	-5.85395
300001	-5.84648
350001	-5.83824
400001	-5.82713
450001	-5.82181
500001	-5.81761
550001	-5.80893
600001	-5.80444
650001	-5.79809
700001	-5.79229
750001	-5.79101
800001	-5.78563
850001	-5.78235
900001	-5.78048
950001	-5.77479
1000001	-5.77312
1050001	-5.77277
1100001	-5.77006
1150001	-5.76876
1200001	-5.76775
1250001	-5.76494
1300001	-5.76448
1350001	-5.76256
1400001	-5.76169
1450001	-5.76137
1500001	-5.76057
1550001	-5.75981
1600001	-5.7578
1650001	-5.75204
1700001	-5.7501
1750001	-5.74887
1800001	-5.74825
1850001	-5.74794
1900001	-5.74737
1950001	-5.74742
2000001	-5.74637
2050001	-5.74724
2100001	-5.74545
2150001	-5.74563
2200001	-5.74485
2250001	-5.74413
2300001	-5.74381
2350001	-5.74452
2400001	-5.74407
2450001	-5.74529
2500001	-5.74391
2550001	-5.7436
2600001	-5.7425
2650001	-5.74073
2700001	-5.74063
2750001	-5.74114
2800001	-5.74044
2850001	-5.73845
2900001	-5.73826
2950001	-5.7373
3000001	-5.73646
3050001	-5.73635
3100001	-5.73637
3150001	-5.73542
3200001	-5.73554
3250001	-5.73561
3300001	-5.7359
3350001	-5.735
3400001	-5.73443
3450001	-5.73509
3500001	-5.7352
3550001	-5.73446
3600001	-5.73501
3650001	-5.73482
3700001	-5.73418
3750001	-5.73319
3800001	-5.73306
3850001	-5.7334
3900001	-5.7322
3950001	-5.73261
4000001	-5.73294
4050001	-5.73239
4100001	-5.73261
4150001	-5.73285
4200001	-5.73185
4250001	-5.73219
4300001	-5.73273
4350001	-5.73245
4400001	-5.73182
4450001	-5.7314
4500001	-5.73201
4550001	-5.73161
4600001	-5.73192
4650001	-5.7313
4700001	-5.73168
4750001	-5.73229
4800001	-5.73127
4850001	-5.73008
4900001	-5.73041
4950001	-5.73011
5000001	-5.73035
5050001	-5.73076
5100001	-5.73059
5150001	-5.73143
5200001	-5.73025
5250001	-5.73065
5300001	-5.72986
5350001	-5.72968
5400001	-5.72968
5450001	-5.73012
5500001	-5.7291
5550001	-5.72897
5600001	-5.72918
5650001	-5.72934
5700001	-5.72947
5750001	-5.72843
5800001	-5.72975
5850001	-5.73007
5900001	-5.72955
5950001	-5.7287
6000001	-5.72939
6050001	-5.72818
6100001	-5.72814
6150001	-5.72843
6200001	-5.72844
6250001	-5.7289
6300001	-5.72901
6350001	-5.72899
6400001	-5.72775
6450001	-5.72763
6500001	-5.72808
6550001	-5.72838
6600001	-5.72838
6650001	-5.72724
6700001	-5.72738
6750001	-5.72695
6800001	-5.7279
6850001	-5.72765
6900001	-5.72801
6950001	-5.72752
7000001	-5.72722
7050001	-5.72688
7100001	-5.72697
7150001	-5.72694
7200001	-5.72664
7250001	-5.72629
7300001	-5.72735
7350001	-5.72658
7400001	-5.72694
7450001	-5.72664
7500001	-5.72665
7550001	-5.72714
7600001	-5.72621
7650001	-5.72725
7700001	-5.72747
7750001	-5.72747
7800001	-5.72703
7850001	-5.72671
7900001	-5.72583
7950001	-5.72637
8000001	-5.72579
8050001	-5.72547
8100001	-5.72561
8150001	-5.72585
8200001	-5.72513
8250001	-5.72483
8300001	-5.72519
8350001	-5.72563
8400001	-5.72476
8450001	-5.72467
8500001	-5.72494
8550001	-5.72495
8600001	-5.72517
8650001	-5.72461
8700001	-5.72509
8750001	-5.72445
8800001	-5.7246
8850001	-5.724
8900001	-5.72326
8950001	-5.72348
};
\addlegendentry{K=20, OBCTR}

\end{axis}
\end{tikzpicture} 

%% file: parameter_vr.tex
\begin{tikzpicture}
  \begin{axis} [
    xlabel = {$\sigma_\epsilon$},
    ylabel = {RMSE},
    xmode = log,
     legend style={legend pos=north east,font=\tiny},
    log basis x={2}
  ]
	\addplot [color=blue, mark=*, thick] table{
0.125	0.907549
0.25	0.880149
0.5	0.875379
1	0.87706
2	0.880122
4	0.88084
8	0.884747
16	0.888974
32	0.891539
64	0.892589
128	0.892589
256	0.890571

    };
    \addlegendentry{$\sigma_r = 1$}

    \addplot [color=purple, mark=*, thick] table{
0.125	0.947333
0.25	0.907227
0.5	0.882739
1	0.874355
2	0.873218
4	0.880131
8	0.8795
16	0.884214
32	0.888056
64	0.890032
128	0.891012
256	0.89153

    };
    \addlegendentry{$\sigma_r = 2$}

    \addplot [color=orange, mark=*, thick] table{
0.125	0.966374
0.25	0.956826
0.5	0.911262
1	0.884274
2	0.876486
4	0.871584
8	0.876469
16	0.87779
32	0.882845
64	0.886237
128	0.888413
256	0.890251

    };
    \addlegendentry{$\sigma_r = 4$}

        \addplot [color=cyan, mark=*, thick] table{
1	0.96927
2	0.916035
4	0.886206
8	0.876943
16	0.877786
32	0.877263
64	0.881147
128	0.884713
256	0.88692

    };
    \addlegendentry{$\sigma_r = 8$}

        \addplot [color=olive, mark=*, thick] table{
8	0.948907
16	0.895442
32	0.881974
64	0.877175
128	0.879294
256	0.883809

    };
    \addlegendentry{$\sigma_r = 16$}

	\end{axis}
\end{tikzpicture} 

%% file: parameter_k.tex
\begin{tikzpicture}
\pgfplotsset{
    scale only axis,
    xmin=0, xmax=100,
}

\begin{axis}[
  axis y line*=left,
  ymin=-6.5, ymax=-5.5,
  xlabel=K,
  ylabel=Held-out likelihood,
  legend style={legend pos=south east}
]
\addplot[color=cyan, mark=square*, thick]
  coordinates{
	(5,	-6.1014)
	(10,	-5.957223)
	(20,-5.80998125)
	(50,	-5.78793)
	(100,	-5.71946)
}; \label{plot_one}
\addlegendentry{Held-out likelihood}
\end{axis}

\begin{axis}[
  axis y line*=right,
  axis x line=none,
  ymin=0.8, ymax=0.9,
  ylabel=RMSE,
  legend style={legend pos=south east}
]
\addplot[color=blue,mark=*,thick, forget plot]
  coordinates{
(5,	0.876323)
(10,	0.873433)
(20,	0.873122)
(50,	0.870256)
(100,	0.865095)
}; \label{plot_two}
\addlegendentry{RMSE}
\addlegendimage{/pgfplots/refstyle=plot_one}\addlegendentry{Held-out likelihood}
\addlegendimage{/pgfplots/refstyle=plot_two}\addlegendentry{plot 2}

\end{axis}

\end{tikzpicture} 